%% file: main.tex
\documentclass{article} 
\usepackage[main, final]{neurips_2026}

\usepackage[utf8]{inputenc} 
\usepackage[T1]{fontenc}    
\usepackage{graphicx}
\usepackage{booktabs}       
\usepackage{amsfonts}
\usepackage{amsmath}
\usepackage{amsthm}
\usepackage{enumitem}
\usepackage{nicefrac}       
\usepackage{microtype}      
\usepackage{xcolor}         
\usepackage{algorithm}
\usepackage{algpseudocode}
\usepackage{makecell}
\usepackage{multirow}
\usepackage{array}
\usepackage{subcaption}
\usepackage{titletoc}
\usepackage{xspace}
\usepackage[table,dvipsnames]{xcolor}
\usepackage{hyperref}
\usepackage{url}
\usepackage[capitalize,noabbrev,nameinlink]{cleveref}
\usepackage[textsize=tiny]{todonotes}

\allowdisplaybreaks

\newcommand{\name}{V-pretraining\xspace}

\theoremstyle{plain}
\newtheorem{theorem}{Theorem}[section]
\newtheorem{proposition}[theorem]{Proposition}

\theoremstyle{definition}

\theoremstyle{remark}

\title{Value-Based Pre-Training With Downstream Feedback: Meta-Learn Your Pre-Training Objectives}

\author{Shuqi Ke\\
Department of ECE\\
Carnegie Mellon University\\
Pittsburgh, PA 15213, USA \\
\texttt{shuqik@andrew.cmu.edu} \\
\And
Giulia Fanti \\
Department of ECE\\
Carnegie Mellon University\\
Pittsburgh, PA 15213, USA \\
\texttt{gfanti@andrew.cmu.edu}
}

\begin{document}

\maketitle

\begin{abstract}
Continued pretraining is typically optimized with a fixed self-supervised task, yet selected and justified by downstream performance. This creates a coarse feedback loop: practitioners evaluate checkpoints, revise data mixtures or objectives, and restart pretraining runs, while individual pretraining updates remain blind to whether they help the capabilities of interest. We ask whether a small set of verifiable downstream examples can provide step-level feedback during continued pretraining without direct learner supervision. $\;$We introduce \name, which separates a \emph{learner} trained only by a self-supervised loss from a lightweight
\emph{task designer} that constructs targets or views for unlabeled batches.
Given the current learner and an unlabeled batch,
\name estimates the downstream value of a candidate target or view construction. It does this by predicting the first-order drop in downstream loss caused by a self-supervised update. 
The designer is trained to increase this value; the learner then applies the resulting self-supervised update, with targets or views detached,  so downstream labels never directly update learner parameters.
We instantiate \name  to learn adaptive top-$K$ soft targets for language modeling and learned views or masks for self-supervised vision. We observe that \name can significantly improve target capabilities without harming generalization on vision and language modalities. For instance, under wall-clock-matched continued pretraining, \name improves GSM8K Pass@1 for Qwen models using 1,024 GSM8K examples only as feedback, including a \(+7.4\) point single-run gain for Qwen2.5-0.5B. 
In vision, \name improves DINOv3 transfer to ADE20K semantic segmentation and NYUv2 depth estimation while preserving ImageNet linear accuracy, indicating that feedback-guided task construction improves target downstream capabilities without collapsing general-purpose representations.
\end{abstract}

\input{section/introduction}
\input{section/vpretraining}
\input{section/experiments}
\input{section/discussion}



\begin{ack}
This work was supported in part by NSF grant CCF-2338772. This work used the DeltaAI system at the National Center for Supercomputing Applications (award OAC 2320345) \citep{bode2025deltaai} through allocation CIS260343 from the Advanced Cyberinfrastructure Coordination Ecosystem: Services \& Support (ACCESS) program \citep{boerner2023access}, which is supported by U.S. National Science Foundation grants 2138259, 2138286, 2138307, 2137603, and 2138296. The authors declare no competing interests.
\end{ack}

\nocite{rathi2026shaping}

\bibliographystyle{plain}
\bibliography{iclr2026_conference}

\newpage
\appendix
\startcontents[appendix]
\printcontents[appendix]{l}{1}{\section*{Appendix Contents}}
\newpage
\crefalias{section}{appendix}
\crefalias{subsection}{appendix}

\input{section/app_algorithm}
\input{section/app_experiments}
\input{section/app_relatedwork}
\input{section/app_multitask}
\input{section/app_main_lang}
\input{section/app_main_vision}
\input{section/app_declare}
\newpage

\end{document}

%% file: section/introduction.tex
\section{Introduction}\label{sec:introduction}

\begin{figure}[t]
    \centering
    \includegraphics[width=\linewidth]{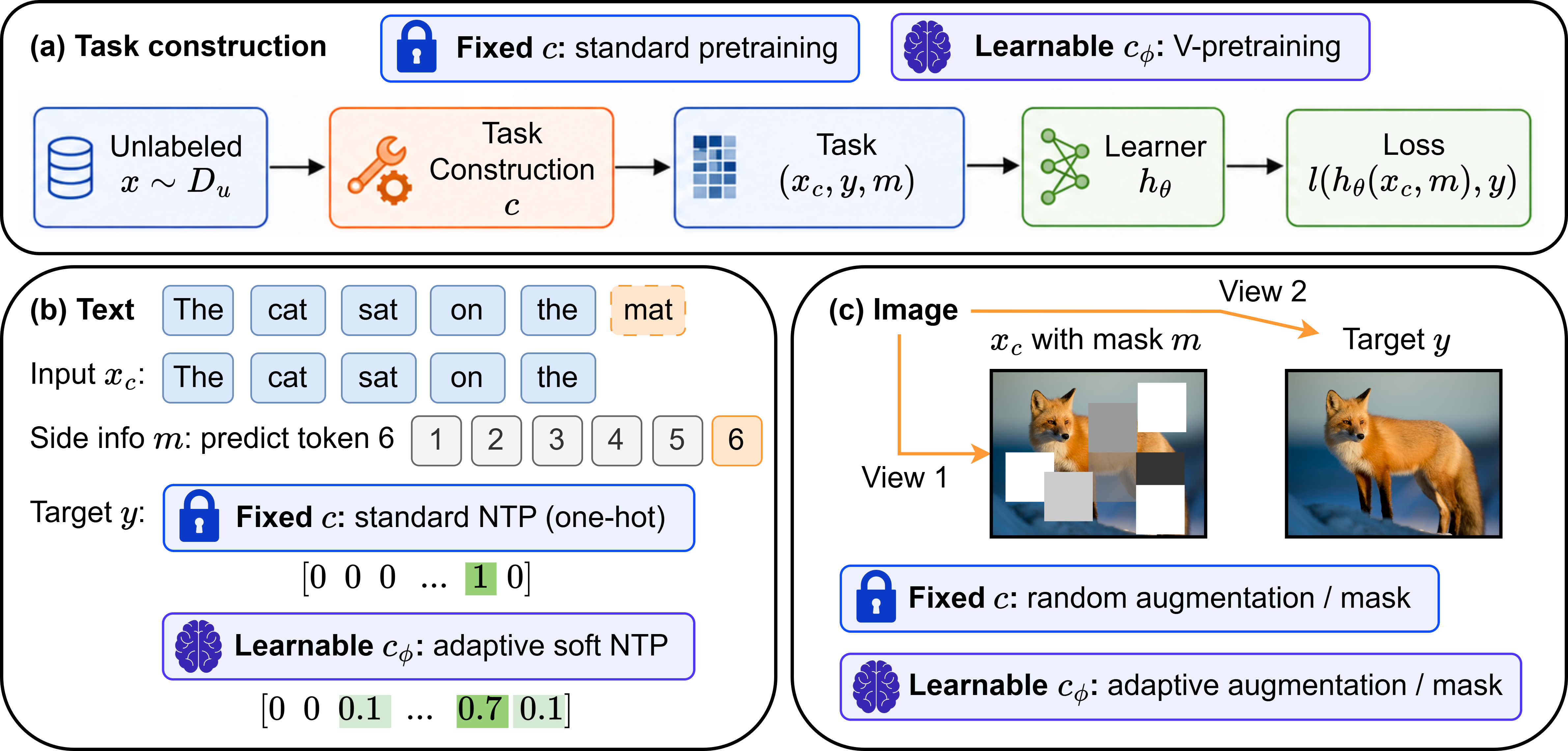}
    \caption{Task construction as the control surface in continued pretraining. A construction rule \(c\) maps each unlabeled example \(x\sim\mathcal D_u\) into a self-supervised prediction problem \((x_c,y,m)\), such as one-hot next-token prediction in language modeling or paired views and targets in DINO-style vision SSL. Standard continued pretraining fixes this rule before training, whereas \name replaces it with a feedback-trained designer \(c_\phi\) while keeping the learner update self-supervised.
    }
    \label{fig:illustration-of-v-pretraining}
\end{figure}

Continued pretraining is a standard way to adapt foundation models to new domains and capabilities \citep{gururangan2020dont,gupta2023continual,ou2025llms}. 
For instance, a language model may be further trained on mathematical text to improve reasoning \citep{lu2025mathcoder}, or a visual backbone may be further trained with self-supervised objectives to improve dense prediction \citep{simeoni2025dinov3}. In both cases, the training signal is usually a fixed proxy task \cite{lecun2016predictive}: next-token prediction for text \citep{gpt3,gpt4,qwen3}, reconstruction/view-based or joint-embedding objectives for images \citep{chen2020simclr,assran2023self,simeoni2025dinov3}. These proxy tasks scale because they do not require dense human annotation \citep{he2022mae,gpt3}. Yet a pretraining run is not judged by the proxy loss itself. It is judged by downstream performance.


This creates a mismatch between the unit of optimization and the unit of selection. Optimization happens one self-supervised update at a time; selection happens after entire continued-pretraining runs are evaluated downstream \citep{morningstar2024augment}. In current practice, downstream feedback enters through a coarse outer loop: practitioners evaluate checkpoints, revise the corpus, objective, curriculum, or augmentation recipe, and launch another run. This makes task-recipe design an expensive black-box search over full pretraining trajectories.


This motivates our question of interest: given a pretrained foundation model, an unlabeled stream for continued pretraining, and a small set of verifiable examples that specify desired downstream capabilities, can we use this
feedback during continued pretraining to improve desired capabilities under a limited compute budget, while preserving the generalization capabilities of the pretrained model?


A natural first-pass solution is to fine-tune on the feedback examples. However, fine-tuning is not necessarily the most efficient use of downstream samples when the feedback set is small. SFT \citep{wei2022finetuned}, preference optimization \citep{rafailov2024dpo}, and RL \citep{ouyang22instruct,shao2024deepseekmath,ahmadian2024back} use downstream examples as learner supervision: labels, preferences, or rewards directly define the learner update. These methods are powerful and complementary, but they answer a different question: how to update the learner on downstream supervision. They do not tell us which unlabeled self-supervised pretraining update was worth taking during a continued-pretraining run; they replace that update with a downstream supervised or reward-based one. Our goal is instead to use downstream examples as a value signal over candidate pretraining updates, without applying downstream gradients or rewards as learner updates.


We introduce \textbf{\name: value-based continued pretraining with downstream feedback}, a framework for the downstream-guided design of self-supervised pretraining tasks (\cref{fig:illustration-of-v-pretraining}). \name complements the target learner, with parameters $\theta$, with a lightweight task designer, with parameters $\phi$. The learner is trained only by a self-supervised pretraining loss. The designer instead dynamically controls how the self-supervised task is constructed during continued pretraining. 
For example, in language modeling, it shapes the next-token target distribution over a small candidate set; in vision, it selects instance-wise views or masks for self-supervised learning. A feedback batch defines a downstream evaluator gradient, but that gradient is not used to update the learner's parameters directly; rather, it is used only to train the task designer.

The primary technical challenge is that the ideal designer objective is long-horizon and bilevel: minimize a downstream task loss function $L_{\rm down}$, i.e. choose task constructions whose induced learner trajectory performs well downstream \citep{maclaurin2015gradient,franceschi2018bilevel}. This is computationally infeasible at pretraining scale. So \name replaces this objective with a one-step value estimate \citep{koh2017understanding,pruthi2020estimating}. If a proposed pretraining task induces gradient $g_{\rm pre}(\theta;\phi)$ and the feedback set induces downstream gradient $g_{\rm down}(\theta)$, then for a small enough $\eta>0$
\begin{equation}
    L_{\rm down}(\theta-\eta g_{\rm pre}) \approx L_{\rm down}(\theta)-\eta g_{\rm down}^{\top}g_{\rm pre}.
\end{equation}
Thus $g_{\rm down}^{\top}g_{\rm pre}$ estimates whether the unlabeled update would reduce downstream loss. \name trains the designer to maximize this value, then detaches the resulting targets or views and updates the learner with the usual pretraining loss. The pretraining loss has a target selected based on the output of the task designer. We instantiate \name in both language and vision modalities. For language models, the designer learns adaptive top-$K$ soft targets for continued next-token pretraining. For vision backbones, it learns instance-wise views for DINO-style self-supervised learning. In both cases, \name changes the task construction, not the learner architecture, optimizer, unlabeled stream, or downstream evaluation protocol.

\paragraph{Contributions.}
We make the following contributions. 
\begin{itemize}[leftmargin=*,topsep=2pt,itemsep=2pt]
    \item \textbf{Conceptual:} We propose \name, a novel framework for continued pretraining. Notably, we formulate downstream-guided continued pretraining as learnable task construction, which introduces a new channel for guidance during continued pretraining.
    This guidance comes without directly supervising the pretraining task, but implicitly injects supervision via the task designer. 
    \item \textbf{Algorithmic:} We propose a 
    step-level objective for \name that can be easily and efficiently optimized, and instantiated for different modalities.
    In particular, we instantiate the same principle as adaptive top-\(K\) target construction for language pretraining and learned view construction for DINO/I-JEPA style vision SSL. In both cases, feedback changes the self-supervised task, not the learner objective.
    \item \textbf{Empirical:} We conduct extensive compute-matched experiments to demonstrate the utility of \name.
    We demonstrate on multiple modalities that \name improves downstream performance on target tasks without harming generalization. Under wall-clock-matched continued pretraining, \name achieves a $+7.4$ point gain or $33\%$ relative improvement on GSM8K Pass@1 in language. In vision, \name improves DINOv3-ViT-L transfer from $51.33$ to $52.40$ mIoU on ADE20K and reduces NYUv2 RMSE from $0.5752$ to $0.5522$. Ablations, one-step probes, decontamination, transfer checks, multitask feedback, and overhead measurements show that the gains require task-relevant feedback and are not explained by generic smoothing, self-distillation, contamination, or direct post-training feedback.
\end{itemize}


%% file: section/vpretraining.tex
\section{\name Framework}


Self-supervised pretraining and continued pretraining do not only specify a loss; they choose how an unlabeled sample is turned into a prediction problem \citep{lecun2016predictive}. We call this choice the \emph{task construction rule}. Given an unlabeled example $x\sim\mathcal{D}_u$ from the pretraining data distribution $\mathcal D_u$, a (possibly randomized) task construction rule $c(\cdot)$ produces a modified input $x_c$, a target $y$, and optional side information $m$, i.e., $(x_c,y,m)\sim c(x)$. The learner predicts $y$ by outputting $\hat y = h_\theta(x_c,m)$ and minimizes
\begin{equation}
    L_{\rm pre}(\theta;c)=\mathbb{E}_{x\sim\mathcal{D}_u}\mathbb{E}_{(x_c,y,m)\sim c(x)}\ell(h_\theta(x_c,m),y).
\end{equation}
In next-token pretraining \citep{gpt3}, given a token sequence $x=(w_1,\ldots,w_n)$, $c$ selects a position $t\in\{1,\dots,n\}$, sets $x_c=w_{<t}$, uses side information $m=t$, and sets the target to the one-hot next token $y=\delta_{w_t}$. In vision masked modeling \citep{he2022mae,xie2021simmim,assran2023self}, $c$ chooses which tokens or patches to hide and defines the reconstruction targets. In DINO-style \citep{simeoni2025dinov3} vision SSL, $c$ samples crops \citep{caron2020multicrop}, augmentations, and masks; the target is the teacher representation induced by the paired view, and the learner predicts it from the student view.

Standard continued pretraining fixes $c$ before the run. This does not mean the constructed tasks are deterministic: token positions, masks, crops, and augmentations can still be random \citep{caron2021emerging,he2022mae}. It means the distribution that constructs inputs and targets is chosen in advance and does not depend on downstream feedback during training. Our language baseline fixes $c$ to one-hot next-token prediction on the continued-pretraining text stream. Our vision baseline fixes $c$ to the DINO view-generation pipeline. Downstream performance may motivate these recipes, but once training starts, each learner update is optimized for the fixed proxy task rather than for an explicit estimate of downstream value.

\name alters the construction rule $c$ and define $c$ as a function of the unlabeled data batch (\cref{fig:illustration-of-v-pretraining}). It keeps the learner architecture, optimizer, loss family, and unlabeled data stream fixed, but replaces the fixed rule $c$ with a designer-controlled rule $c_\phi$. In natural language, $c_\phi$ changes the target distribution for next-token prediction while keeping the context and text stream fixed. In vision, $c_\phi$ changes instance-wise views or masks while keeping the SSL objective fixed. Thus the controlled object is not the learner itself, but the self-supervised task that generates the learner's next update.

\subsection{Indirect feedback through task construction}

Let $\mathcal{D}_{\rm fb}$ be a small feedback set with labels, rewards, or other verifiable downstream signals, and let $\mathcal{D}_{\rm eval}$ be a held-out evaluation set. Direct post-training methods use $\mathcal{D}_{\rm fb}$ as learner supervision: SFT \citep{ouyang22instruct,wei2022finetuned}, preference optimization \citep{meng2024simpo,rafailov2024dpo}, or RL \citep{ahmadian2024back,shao2024deepseekmath} supply a downstream loss or reward update to $\theta$. \name uses the same kind of information during pretraining phase. The feedback set may train the task designer but we do \emph{not} use it to directly train the learner.

Formally, every learner update must remain a self-supervised pretraining update on an unlabeled batch $B^u_t\subset\mathcal{D}_u$:
\begin{equation}
    g^{\rm learn}_t=
    \nabla_\theta L_{\rm pre}\bigl(\theta_t;\operatorname{sg}(c_{\phi_t}(B^u_t))\bigr),
    \qquad
    \theta_{t+1}=\theta_t-\eta_t g^{\rm learn}_t.
\end{equation}
Here $\operatorname{sg}(\cdot)$ denotes stop-gradient through the constructed targets or views. We write parameter updates using gradient descent for clarity; other optimizers like AdamW \citep{loshchilov2018decoupled} can simply replace $-\eta_t g_t$ with the adaptive optimizer step, and all comparisons use the same learner optimizer and schedule. The key invariant is the gradient supplied to the learner: no term of the form $\nabla_\theta L_{\rm down}(\theta_t;\mathcal{D}_{\rm fb})$ is applied to $\theta$.

This distinction is essential. Fine-tuning answers how the learner should fit the feedback examples \citep{wei2022finetuned}. \name asks which unlabeled self-supervised update is worth taking before the run is over (\cref{fig:illustration-of-v-pretraining}). The feedback signal is therefore direct-to-designer but indirect-to-learner: it can change $c_\phi(B^u_t)$, and hence the pretraining gradient, but it cannot replace that gradient with a downstream supervised or reward gradient.

\subsection{A local value objective for task design}

The ideal designer would choose a construction rule whose full continued-pretraining trajectory performs well downstream:
\begin{align}
\min_\phi L_{\rm down}&(\theta_T(\phi)),\\
\theta_{t+1}(\phi)=\theta_t(\phi)-\eta_t\nabla_\theta L_{\rm pre}&\bigl(\theta_t(\phi);c_\phi(B^u_t)\bigr),\quad t=0,\ldots,T-1.
\end{align}
Differentiating through this trajectory is impractical at pretraining scale \citep{finn2017adax,rajeswaran2019meta,franceschi2018bilevel,ji2021bilevel}. \name instead asks a local question at each step: among the tasks that can be constructed from the current unlabeled batch, which one induces a learner gradient that would most decrease the feedback loss?
For an unlabeled batch $B^u$ and a labeled feedback batch $B^f$, the designer induces a candidate pretraining gradient
\begin{equation}
g_{\rm pre}(\theta;\phi,B^u)=\nabla_\theta L_{\rm pre}\bigl(\theta;c_\phi(B^u)\bigr),
\end{equation}
and the feedback batch defines an evaluator gradient
\begin{equation}
g_{\rm down}(\theta;B^f)=\nabla_\theta L_{\rm down}(\theta;B^f).
\end{equation}
The candidate learner update is $\theta^+=\theta-\eta g_{\rm pre}$. A first-order expansion \citep{koh2017understanding,pruthi2020estimating} gives
\begin{equation}
L_{\rm down}(\theta^+;B^f)\approx L_{\rm down}(\theta;B^f)-\eta g_{\rm down}(\theta;B^f)^\top g_{\rm pre}(\theta;\phi,B^u).
\end{equation}
We therefore define the value of the proposed pretraining update as the inner product
\begin{equation}
V(\phi;\theta,B^u,B^f)=g_{\rm down}(\theta;B^f)^\top g_{\rm pre}(\theta;\phi,B^u),
\end{equation}
and train the designer with $L_{\rm meta}(\phi)=-V(\phi;\theta,B^u,B^f)$. Gradients through $g_{\rm down}$ are stopped; the meta-gradient changes $\phi$ according to how its task construction changes the learner gradient. For efficiency, we compute the dot product on a parameter subset $S$ such as the last transformer blocks or task-relevant projection layers. After the designer update, the task is reconstructed and detached, and the learner applies the gradient-descent pretraining update above. Under smoothness, maximizing $V$ maximizes the first-order term in a lower bound on one-step downstream improvement; the proofs and full algorithm are in \cref{app:algorithm} and \cref{app:value-objective}.

\subsection{One-step justification}
\label{sec:one_step_justification}

We now state the local guarantee underlying the value objective. The result is not a long-horizon convergence theorem. Instead, it shows that the alignment objective maximizes a first-order lower bound on one-step downstream improvement.

\begin{proposition}[1-step downstream improvement lower-bound]
Let $L_{\mathrm{down}}$ be $L$-smooth in $\theta$. Let
\begin{equation}
    \theta^{+}
    =
    \theta
    -
    \eta g_{\mathrm{pre}}(\theta;\phi)
\end{equation}
for step size $\eta>0$, and define
\begin{equation}
    V(\phi;\theta)
    =
    g_{\mathrm{down}}(\theta)^{\top}
    g_{\mathrm{pre}}(\theta;\phi),
    \qquad
    g_{\mathrm{down}}(\theta)=\nabla_{\theta}L_{\mathrm{down}}(\theta).
\end{equation}
Then
\begin{equation}
    L_{\mathrm{down}}(\theta)
    -
    L_{\mathrm{down}}(\theta^{+})
    \ge
    \eta V(\phi;\theta)
    -
    \frac{L\eta^2}{2}
    \left\|g_{\mathrm{pre}}(\theta;\phi)\right\|_2^2.
\end{equation}
\end{proposition}

The proposition explains the role of $V$. For sufficiently small learner steps, or when the pretraining update norm is controlled, increasing $g_{\mathrm{down}}^{\top}g_{\mathrm{pre}}$ increases a certified lower bound on one-step downstream loss decrease. Equivalently, define the one-step bilevel objective
\begin{equation}
    J(\phi;\theta)
    =
    L_{\mathrm{down}}
    \left(
        \theta
        -
        \eta\nabla_{\theta}L_{\mathrm{pre}}(\theta;\phi)
    \right).
\end{equation}
A Taylor expansion gives $J(\phi;\theta)=L_{\mathrm{down}}(\theta)-\eta V(\phi;\theta)+O(\eta^2)$. Thus maximizing $V$ is equivalent to minimizing the first-order approximation of the one-step downstream objective.

This local result is deliberately modest. It does not claim that every high-value step guarantees monotonic downstream improvement over a long training trajectory. Stochastic gradients, optimizer state, distribution shift between feedback and evaluation, and interactions across steps can all affect the final trajectory. The purpose of the value objective is to provide a scalable online signal for task design. \cref{sec:experiments} evaluates whether it improves downstream performance over full continued-pretraining runs. We leave the detailed proofs to \cref{app:proofs}.

\subsection{Instantiations}

\textbf{Language: adaptive top-$K$ soft targets.} For a token sequence $x=(w_1,w_2,\ldots)$, standard next-token pretraining uses context $w_{<t}$ and one-hot target $y=\delta_{w_t}$
\name keeps the context, token positions, and text stream fixed, but lets the designer replace the one-hot target with a bounded soft target over a small candidate set $C_t$:
\begin{equation}
    C_t=\{w_t\}\cup\operatorname{TopK}_{K-1}\bigl(\operatorname{sg}(p_\theta(\cdot\mid w_{<t}))\setminus\{w_t\}\bigr).
\end{equation}
The true next token is always included. The task construction rule $c_\phi$ outputs a distribution $r_{\phi,t}\in\Delta(C_t)$ and a gating constant $\alpha_{\phi,t}\in[0,\alpha_{\max}]$, producing weights for each element  $v\in C_t$:
\begin{equation}
    q_{\phi,t}(v)=(1-\alpha_{\phi,t})\mathbf{1}[v=w_t]+\alpha_{\phi,t}r_{\phi,t}(v),\qquad v\in C_t.
\end{equation}
The learner minimizes cross-entropy to detached $q_{\phi,t}$ on the continued-pretraining text. Feedback examples such as GSM8K \citep{cobbe2021gsm8k} problems define $L_{\rm down}$ for the designer, but they are not inserted as supervised learner examples.

\textbf{Vision: learned views for self-supervised learning.} We take DINO-style \citep{caron2021emerging,oquab2023dinov2,simeoni2025dinov3} continued SSL as an example here. The base pipeline samples views $V_0(x)=\{v^0_1,\ldots,v^0_M\}$ for an unlabeled image. \name keeps the DINO loss and student--teacher learner fixed, but lets the designer modify selected views through instance-wise masks or view parameters, producing $V_\phi(x)$. The learner uses the same SSL loss as the baseline,
\begin{equation}
    L_{\rm pre}^{\rm vis}(\theta;\phi,B^u)=\frac{1}{|B^u|}\sum_{x\in B^u}\ell_{\rm SSL}(\theta,\bar\theta;V_\phi(x)),
\end{equation}
where $\bar\theta$ is the EMA teacher used to anchor the DINO SSL pipeline. The feedback evaluator uses lightweight downstream task heads like segmentation and depth estimation on top of the backbone,
\begin{equation}
    L_{\rm down}^{\rm vis}=\omega_{\rm seg}L_{\rm seg}(H_{\rm seg}(F_\theta(x)),y_{\rm seg})+
    \omega_{\rm dep}L_{\rm dep}(H_{\rm dep}(F_\theta(x)),y_{\rm dep})+\cdots,
\end{equation}
Segmentation \citep{zhou2017ade20k} or depth estimation \citep{silberman2012nyu} labels train the task designer through $g_{\rm down}$; the backbone update remains the detached SSL loss on unlabeled ImageNet \citep{deng2009imagenet} images. By simply replacing the task construction rule $c$, this instantiation also applies to other SSL method backends like I-JEPA \citep{assran2023self}.

%% file: section/experiments.tex
\section{Experiments}\label{sec:experiments}

We evaluate whether downstream feedback can steer continued pretraining through task construction rather than direct learner supervision. The main experiments ask: (i) Does \name improve target capabilities under a strict compute comparison? (ii) Does feedback steering over-specialize the model to the feedback task and harm broader transfer? (iii) Are the gains caused by leakage or shortcut
supervision, where the learner effectively sees benchmark or feedback labels? (iv) Does the gradient-alignment value signal matter, or are the gains explained by generic smoothing, self-distillation, or additional stochasticity? (v) How does indirect feedback compare with direct post-training, and what overhead does it introduce? (vi) Can the same feedback channel scale across model sizes and support multitask downstream feedback?

\subsection{Protocol}

All main comparisons are matched by \textbf{wall-clock training time on the same hardware configuration}. For each model size, modality, and method, training is run for the same elapsed time under the same hardware and software configuration \citep{klein2019hp,mattson2020mlperf,kumar2021limits}. The initial learner, unlabeled stream, sequence length or crop configuration, optimizer, numerical precision, and data-loading protocol are held fixed. If \name introduces additional per-step overhead, it must pay that cost within the same elapsed time. Thus the baseline is allowed to complete as many learner updates and process as many unlabeled tokens or images as its faster training loop permits. \name receives no extra time to compensate for task-designer computation.

\textbf{Language.} We continue pretraining Qwen1.5 \citep{qwen1.5} and Qwen2.5 \citep{qwen2.5} base models on NuminaMath-CoT \citep{numina_math_datasets} in main experiments. The \textbf{baseline} is standard next-token continued pretraining with one-hot targets. \textbf{\name} uses the same learner configuration and text stream, but replaces one-hot targets with designer-shaped top-$K$ targets. The feedback set contains 1,024 GSM8K training examples, used only to compute $g_{\rm down}$ for the designer. Evaluation is GSM8K test Pass@1 with greedy decoding. See \cref{app:main-language-setup} for more details and hyper-parameters.

\textbf{Vision.} We continue self-supervised training of DINOv3 ViT-B and ViT-L backbones \citep{simeoni2025dinov3} on ImageNet-1K \citep{deng2009imagenet} with a DINO-style student-teacher objective. The \textbf{baseline} uses the fixed view-generation pipeline of vanilla DINOv3. \textbf{\name} uses learned instance-wise views with the same SSL objective and learner schedule. Dense-task feedback comes from small labeled ADE20K and NYUv2 pools. We report ADE20K mIoU, NYUv2 RMSE, ImageNet linear accuracy, and transfer diagnostics \citep{silberman2012nyu,zhou2017ade20k,radenovi2018revisit}.

\subsection{\name improves target capabilities}

\cref{tab:main-results} reports the main wall-clock-matched results. In this part, we focus on tasks that are aligned with the capabilities of the small feedback dataset used to train the task designer. In language, \name improves GSM8K Pass@1 across the tested Qwen models. The largest gain is the Qwen2.5-0.5B single-run result, improving from 22.20 to 29.60 (a \textbf{33\% gain}). In the replicated Qwen1.5 runs, \name improves 0.5B, 4B, and 7B models, with the strongest replicated gain at 4B. The gains decrease with model size, suggesting that larger learners already extract more math-relevant signal from standard continued next-token prediction.

In vision, we provide downstream feedback from dense prediction tasks for \name. \name improves dense prediction while preserving global recognition (a task that was \emph{not} used to train the task designer). For DINOv3 ViT-B and ViT-L, ADE20K mIoU increases and NYUv2 RMSE decreases; ImageNet linear accuracy also slightly improves. These results support the feasibility claim: downstream feedback can improve continued pretraining through task construction, even though the learner update remains self-supervised and the training time is matched.

\begin{table}
\centering
\small
\begin{tabular}{llllr}
\toprule
\textbf{Modality} & \textbf{Benchmark} & \textbf{Model} &
\textbf{Baseline} & \textbf{\name} \\
\midrule
\multirow{4}{*}{Language} & MATH Pass@1 $\uparrow$ & Qwen2.5-0.5B &
$22.20$ & $\mathbf{29.6}$ \\
 & GSM8K Pass@1 $\uparrow$ & Qwen1.5-0.5B &
$19.15${\tiny${\pm}1.16$} & $\mathbf{22.67}${\tiny$\mathbf{{\pm}1.05}$} \\
 & GSM8K Pass@1 $\uparrow$ & Qwen1.5-4B &
$56.48${\tiny${\pm}1.56$} & $\mathbf{58.98}${\tiny$\mathbf{{\pm}1.03}$} \\
 & GSM8K Pass@1 $\uparrow$ & Qwen1.5-7B &
$65.26${\tiny${\pm}1.06$} & $\mathbf{66.17}${\tiny$\mathbf{{\pm}0.63}$} \\
\midrule
\multirow{9}{*}{Vision} & NYUv2 RMSE $\downarrow$ & DINOv3-ViT-B &
$0.5888$ & $\mathbf{0.5697}$ \\
 & NYUv2 RMSE $\downarrow$ & DINOv3-ViT-L &
$0.5752$ & $\mathbf{0.5522}$ \\
 & NYUv2 RMSE $\downarrow$ & I-JEPA-ViT-H &
$0.7835$ & $\mathbf{0.7797}$ \\
 & ADE20K mIoU $\uparrow$ & DINOv3-ViT-B &
$48.82$ & $\mathbf{49.68}$ \\
 & ADE20K mIoU $\uparrow$ & DINOv3-ViT-L &
$51.33$ & $\mathbf{52.47}$ \\
 & ADE20K mIoU $\uparrow$ & I-JEPA-ViT-H &
$28.03$ & $\mathbf{28.19}$ \\
 & ImageNet-1K Acc. $\uparrow$ & DINOv3-ViT-B &
$80.74$ & $\mathbf{81.01}$ \\
 & ImageNet-1K Acc. $\uparrow$ & DINOv3-ViT-L &
$84.07$ & $\mathbf{84.59}$ \\
 & ImageNet-1K Acc. $\uparrow$ & I-JEPA-ViT-H &
$92.40$ & $\mathbf{92.50}$ \\
\bottomrule
\end{tabular}
\caption{Main target-capability results under matched wall-clock training budgets. Feedback examples train only the task designer; the learner is updated only by a self-supervised loss on the unlabeled stream. Qwen2.5-0.5B is a single-run diagnostic; Qwen1.5 results report mean and standard deviation.}
\label{tab:main-results}
\end{table}

\subsection{\name avoids over-specialization}

Because the feedback sets are small and task-specific, \name could improve
the target metric by steering the learner too narrowly. We therefore evaluate tasks that are not used as feedback.

\cref{tab:generalization_language} evaluates whether feedback steering harms tasks not used as feedback. For language, the OMEGA benchmark \citep{sun2025omega} measures math reasoning under distribution shift, while MMLU benchmark \citep{hendrycks2021mmlu} measures broader zero-shot multiple-choice knowledge and reasoning. \name improves OMEGA for the 4B model, remains similar for 0.5B and 7B, and leaves MMLU nearly unchanged for 4B and 7B. The 0.5B model, however, drops in MMLU. This indicates that \name is a steering mechanism, not a guarantee of monotonic improvement on every unrelated task; small learners can be more sensitive to narrow feedback.

For vision, we evaluate frozen ViT-L representations on Oxford5k \& Paris6k instance retrieval \citep{radenovi2018revisit}, which are not used as feedback tasks. \cref{tab:generalization_vision} shows that dense-task feedback improves most retrieval protocols, including Oxford Easy/Medium and all Paris protocols. Oxford Hard is slightly lower. Overall, dense feedback does not collapse the representation onto merely segmentation or depth, though improvements are not uniform across all retrieval settings.

\begin{table}
    \centering
    \begin{minipage}[t]{0.46\textwidth}
        \centering
        \resizebox{\linewidth}{!}{%
        \begin{tabular}{llcc}
        \toprule
        \textbf{Benchmark} & \textbf{Model} & \textbf{Baseline} & \textbf{\name} \\
        \midrule
        OMEGA Acc. $\uparrow$ & 0.5B & $0.65$ & $0.65$ \\
        OMEGA Acc. $\uparrow$ & 4B & $1.44$ & $\mathbf{1.88}$ \\
        OMEGA Acc. $\uparrow$ & 7B & $\mathbf{1.52}$ & $1.50$ \\
        \midrule
        MMLU Acc. $\uparrow$ & 0.5B & $\mathbf{38.08}$ & $35.01$ \\
        MMLU Acc. $\uparrow$ & 4B & $53.32$ & $\mathbf{53.51}$ \\
        MMLU Acc. $\uparrow$ & 7B & $\mathbf{58.81}$ & $58.68$ \\
        \bottomrule
        \end{tabular}
        }
        \caption{Language generalization beyond the GSM8K feedback task. \name does not over-specialize larger models onto a single downstream task, but the 0.5B model shows a generalization drop on MMLU.}
        \label{tab:generalization_language}
    \end{minipage}\hfill 
    \begin{minipage}[t]{0.5\textwidth}
        \centering
        \resizebox{\linewidth}{!}{%
        \begin{tabular}{llcc}
        \toprule
        \textbf{Dataset} & \textbf{Protocol} & \textbf{Baseline} & \textbf{\name} \\
        \midrule
        R-Oxford5k mAP $\uparrow$ & Easy & $0.5268$ & $\mathbf{0.6048}$ \\
        R-Oxford5k mAP $\uparrow$ & Medium & $0.4072$ & $\mathbf{0.4557}$ \\
        R-Oxford5k mAP $\uparrow$ & Hard & $\mathbf{0.0867}$ & $0.0820$ \\
        \midrule
        R-Paris6k mAP $\uparrow$ & Easy & $0.5433$ & $\mathbf{0.5973}$ \\
        R-Paris6k mAP $\uparrow$ & Medium & $0.6332$ & $\mathbf{0.7005}$ \\
        R-Paris6k mAP $\uparrow$ & Hard & $0.2208$ & $\mathbf{0.2509}$ \\
        \bottomrule
        \end{tabular}
        }
        \caption{Vision transfer beyond the dense feedback tasks. We evaluate frozen ViT-L representations on instance retrieval. Dense-task feedback improves most
        retrieval protocols, with a small decrease on Oxford Hard.}
        \label{tab:generalization_vision}
    \end{minipage}
\end{table}

\subsection{Does the task designer create a shortcut for the learner?}

A separate concern is that \name might improve the target metric through a shortcut rather than through better self-supervised updates. There are two possible shortcuts. First, the unlabeled continued-pretraining stream might contain near-duplicates of evaluation examples. Second, the task designer might effectively turn feedback examples into learner supervision. We test the first possibility experimentally and rule out the second by the training channel.

\textbf{Decontamination.} We decontaminate NuminaMath-CoT by removing similar samples of GSM8K and MATH using MinHash LSH and $n$-gram Jaccard similarity \citep{gionis1999similarity,cobbe2021training}. We provide details of this decontamination in \cref{app:decontamination}.
We then retrain the Qwen1.5-4B baseline and \name models under the same learner update budget. \cref{tab:decontam} shows that \name remains above the baseline after decontamination. The margin is smaller than in the original run, but the result suggests that the main gain is not primarily caused by memorizing benchmark-like examples in the unlabeled stream.

\begin{table}
\centering
\small
\begin{tabular}{lcc}
\toprule
\textbf{Unlabeled stream} & \textbf{Baseline} & \textbf{\name} \\
\midrule
Original NuminaMath-CoT & $56.48$ & $\mathbf{58.98}$ \\
Decontaminated NuminaMath-CoT & $56.7$ & $\mathbf{57.5}$ \\
\bottomrule
\end{tabular}
\caption{
GSM8K Pass@1 for Qwen1.5-4B before and after removing near-duplicates of GSM8K and MATH from the continued-pretraining stream. Baseline and \name are compared under the same wall-clock training budget. Decontamination does not remove the advantage of \name.
}
\label{tab:decontam}
\end{table}

\textbf{No shortcut learner supervision.} The second shortcut is ruled out by construction. During a \name learner step, the learner never receives a gradient of the form \(\nabla_\theta L_{\rm down}(\theta;D_{\rm fb})\), and feedback examples are not inserted into the learner's pretraining batch. The designer can change the target distribution in language or the views/masks in vision, but these constructed targets or views are detached before the learner update.

\subsection{Downstream task signal matters for \name}

\begin{table}
    \centering
    \begin{minipage}[t]{0.35\textwidth}
        \centering
        \resizebox{\linewidth}{!}{%
        \begin{tabular}{lc}
        \toprule
        \textbf{Method} & \textbf{GSM8K Pass@1 $\uparrow$} \\
        \midrule
        Baseline & $56.48$\\
        Rand-feedback & $54.31$\\
        Uni-smoothing & $54.58$\\
        Self-distillation & $57.61$\\
        \name & $\mathbf{58.98}$\\
        \bottomrule
        \end{tabular}
        }
        \caption{Ablations in the Qwen1.5-4B setting. \name is the only variant that trains the designer with $g_{\rm down}^{\top}g_{\rm pre}$.}
        \label{tab:ablations}
    \end{minipage}
    \hfill\begin{minipage}[t]{0.61\textwidth}
        \centering
        \resizebox{\linewidth}{!}{%
        \begin{tabular}{lccc}
        \toprule
        \multicolumn{4}{l}{\textit{Vision multitask feedback}} \\
        \textbf{Method} & \textbf{ADE20K mIoU $\uparrow$} & \textbf{NYUv2 RMSE $\downarrow$} & -- \\
        \midrule
        Baseline & 48.85 & 0.4139 & -- \\
        \name & \textbf{49.49} & \textbf{0.4135} & -- \\
        \midrule
        \multicolumn{4}{l}{\textit{Language multitask feedback}} \\
        \textbf{Method} & \textbf{GSM8K Acc. $\uparrow$} & \textbf{MMLU Acc. $\uparrow$} & \textbf{MMBP Acc. $\uparrow$} \\
        \midrule
        Baseline & 44.28 & 41.90 & 4.00 \\
        \name & \textbf{45.26} & \textbf{42.17} & 4.00 \\
        \bottomrule
        \end{tabular}%
        }
        \caption{Multitask feedback under wall-clock matching. Each \name\ row is a single run using a combined feedback gradient rather than separately tuned single-task feedback.}
        \label{tab:multi}
    \end{minipage}
\end{table}

The main results could have several alternative explanations, including generic label smoothing, self-distillation, or additional stochasticity. We therefore isolate the role of downstream-aligned value feedback in the Qwen1.5-4B setting. Random feedback replaces $g_{\rm down}$ with random vectors \citep{cheon2024rand-pretrain}. Uniform top-$K$ smoothing uses soft targets without downstream feedback \citep{garcin2022smooth,peng2025simko}. Self top-$K$ distillation trains on the learner's own candidate distribution \citep{frydenlund2022rank}. \cref{tab:ablations} shows that these alternatives do not match \name under the same wall-clock budget. Random feedback and uniform smoothing fall below the one-hot baseline, and self-distillation improves over baseline but remains below downstream-aligned value feedback. Thus the gain is not explained by soft labels or extra meta-gradient noise; it requires task-relevant feedback.

\subsection{\name as a lower-cost feedback before post-training}

We show that \name can be a low-cost complement to post-training. Direct fine-tuning methods such as SFT \citep{ouyang22instruct,wei2022finetuned} or GRPO \citep{shao2024deepseekmath} use downstream labels or rewards as learner updates. \name uses them only to train the task designer.

\begin{figure}
    \centering
    \includegraphics[width=0.5\linewidth]{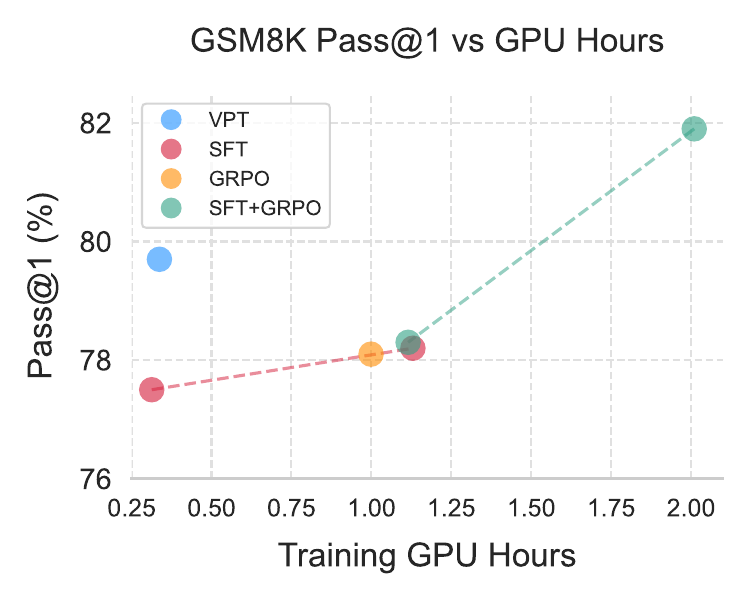}
    \caption{
    Comparing direct and indirect feedback approaches under a shared math data source. SFT and \name train on OpenMathReasoning CoT traces, while GRPO trains on OpenMathReasoning questions. \name uses downstream feedback indirectly through target construction, whereas SFT and GRPO apply supervised or reward feedback directly to the learner. The comparison is diagnostic rather than an apples-to-apples post-training benchmark: SFT+GRPO achieves the highest Pass@1, while \name provides a lower-cost pre-post-training feedback channel.}
    \label{fig:sft-grpo-cost}
\end{figure}

We compare \name with the common direct fine-tuning alternatives under a different shared math data source (\Cref{fig:sft-grpo-cost}) to show that the trend is robust to data distribution. SFT and \name both train on OpenMathReasoning \citep{moshkov2025aimo2}: SFT uses the dataset as direct learner supervision, while \name uses it as the continued training stream with feedback-guided target construction. GRPO also trains on math questions in OpenMathReasoning. So the direct-feedback baselines and \name are exposed to the same source of math problems but use the signal through different training channels. This comparison is therefore a data-source-controlled diagnostic. 

\cref{fig:sft-grpo-cost} shows that \name reaches higher GSM8K Pass@1 than SFT runs with comparable or larger additional GPU hours, and also outperforms GRPO alone in this setting. The SFT+GRPO pipeline achieves the highest Pass@1, but at larger training cost. These results support the intended interpretation: \name can inject useful math capability during continued training before direct post-training, making it a compute-efficient complement to SFT or RL.

\textbf{Runtime overhead.} The preceding comparisons measure downstream performance per reported GPU-hour. We also profile the implementation overhead of the indirect channel itself in a representative language run. This answers a separate question: how much throughput
and memory does \name add relative to standard continued pretraining, before any downstream accuracy is considered?

The \name run uses a lightweight task designer. The value objective aligns the last two learner layers and performs a meta update every 8 learner optimizer steps. Raw optimizer-step time is not the
right cost metric because the meta run uses smaller gradient accumulation to fit the designer and meta-gradient computation. With microbatch size 4 and sequence length 512, the baseline processes \(4\times32\times512=65{,}536\) tokens per optimizer step, while \name processes \(4\times20\times512=40{,}960\) tokens per optimizer step.

\begin{table}
\centering
\small
\setlength{\tabcolsep}{4pt}
\begin{tabular}{lrrrrr}
\toprule
\textbf{Method} & \textbf{Grad. accum.} & \textbf{Tokens/step} & \textbf{Step time} & \textbf{Tokens/s $\uparrow$} & \textbf{Peak memory} \\
\midrule
Baseline & 32 & 65{,}536 & 2.45s & 26.7k & 31{,}302 MiB \\
\name & 20 & 40{,}960 & 1.69s & 24.2k & 36{,}440 MiB \\
\bottomrule
\end{tabular}
\caption{Runtime profiling for a representative current language run. Both methods use microbatch size 4 and sequence length 512. \name uses a smaller gradient-accumulation factor to fit the designer and meta-gradient computation, so throughput rather than raw step time is the primary comparison.}
\label{tab:runtime-overhead}
\end{table}

\cref{tab:runtime-overhead} shows that the configured task designer adds measurable cost. Token throughput decreases by \(9.4\%\), and peak memory increases by \(5{,}138\) MiB, or \(16.4\%\). The shorter optimizer-step time for
\name is therefore not evidence of lower cost, since each optimizer step processes fewer tokens. This overhead is already charged in our wall-clock-matched main results: the baseline is allowed to process more tokens or learner steps within
the same elapsed time. \cref{app:additional-language} reports feedback-set coverage, Pass@\(k\), token-efficiency curves, and additional language diagnostics.

\subsection{Capacity scaling and multitask control}

\cref{tab:main-results}  tested whether the feedback channel transfers across learner scales (measured in number of parameters). A stricter test is whether this channel can compose multiple downstream value
signals in one continued-pretraining run. For feedback tasks
$j=1,\ldots,J$, we use a weighted evaluator gradient
\begin{equation}
    g_{\rm down}=\sum_{j=1}^J \omega_j g_j,
\end{equation}
where $g_j$ is the feedback gradient for task $j$ and $\omega_j$ controls its relative weight. The designer is then trained to construct self-supervised updates whose learner gradients align with this combined value signal.

For the language multitask run, we use a shared continued-pretraining mixture with three domains: math, code, and general instruction following. Each domain contains a raw component and a high-quality instruction component: OpenWebMath
\citep{paster2023openwebmath} and MetaMathQA \citep{yu2023metamath} for math, codeparrot-clean \citep{codeparrot_clean2022} and Magicoder \citep{wei2024magicoder} for code, and C4 \citep{2020t5} and Alpaca \citep{taori2023alpaca} for general text (see \cref{app:multitask-data}). The mixture assigns 70\% probability mass to raw data and 30\% to high-quality instruction data, balanced equally across domains. Both the baseline and \name use the same mixture. Feedback batches are drawn only from the high-quality domain-matched pools and are used to compute the combined evaluator gradient, not to update the learner directly. We present the results in \cref{tab:multi}. \name improves capabilities for both GSM8K ($44.28\rightarrow45.26$) and MMLU ($41.9\rightarrow42.17$). 

%% file: section/discussion.tex
\section{Contributions and Limitations}\label{sec:contribution-limitation}

Our contributions can be summarized as follows. (1) We formulate continued pretraining with downstream feedback as task design rather than learner supervision: a small verifier trains a controller that constructs pretraining targets or views, while the learner remains trained only by a self-supervised loss. (2) We derive a scalable step-level value objective, $g_{\mathrm{down}}^{\top}g_{\mathrm{pre}}$, showing that it is the first-order surrogate for one-step downstream improvement and avoids differentiating through full pretraining trajectories. (3) We instantiate the same principle in language and vision, as adaptive top-K target construction for next-token pretraining and learned view construction for self-supervised vision. (4) We provide compute-matched evidence and controls showing that downstream feedback improves value per learner update and that the effect is not explained by random feedback, fixed smoothing, or self-distillation.

\name provides a local value signal, not a long-horizon optimality guarantee. The alignment objective can be noisy, depends on feedback quality, and may over-steer small learners toward narrow feedback tasks. The method also introduces additional computation for task generation and value-gradient estimation, although our main comparisons match wall-clock time.

%% file: section/app_algorithm.tex
\section{Algorithm}\label{app:algorithm}

\subsection{\name algorithm}

\begin{algorithm}[t]
\caption{\name}
\label{alg:vtraining}
\begin{algorithmic}[1]
\Require Initial learner $\theta_0$, task designer $\phi_0$, unlabeled stream $\mathcal{D}_u$, feedback set $\mathcal{D}_{\mathrm{fb}}$, learner budget $T$, meta-update period $r$, alignment parameter subset $\mathcal{S}$
\For{$t=0,\ldots,T-1$}
    \State Sample unlabeled batch $B_t^u \sim \mathcal{D}_u$.
    \If{$t \bmod r = 0$}
        \State Sample feedback batch $B_t^f \sim \mathcal{D}_{\mathrm{fb}}$.
        \State Construct task $c_{\phi_t}(B_t^u)$ with gradients enabled for $\phi_t$.
        \State Compute $L_{\mathrm{pre}}^{\mathrm{meta}}(\theta_t;\phi_t,B_t^u)$.
        \State $g_{\mathrm{pre},\mathcal{S}} \gets \nabla_{\theta_{\mathcal{S}}} L_{\mathrm{pre}}^{\mathrm{meta}}(\theta_t;\phi_t,B_t^u)$. \Comment{create graph}
        \State $g_{\mathrm{down},\mathcal{S}} \gets \mathrm{sg}\!\left(\nabla_{\theta_{\mathcal{S}}} L_{\mathrm{down}}(\theta_t;B_t^f)\right)$.
        \State $L_{\mathrm{meta}} \gets - g_{\mathrm{down},\mathcal{S}}^{\top} g_{\mathrm{pre},\mathcal{S}}$.
        \State $\phi_{t+\frac{1}{2}} \gets \mathrm{Opt}_{\phi}\!\left(\phi_t,\nabla_{\phi}L_{\mathrm{meta}}\right)$.
    \Else
        \State $\phi_{t+\frac{1}{2}} \gets \phi_t$.
    \EndIf
    \State Construct task $c_{\phi_{t+\frac{1}{2}}}(B_t^u)$ and detach its targets, masks, or views.
    \State $L_{\mathrm{pre}}^{\mathrm{learn}} \gets L_{\mathrm{pre}}\!\left(\theta_t;\mathrm{sg}\!\left(c_{\phi_{t+\frac{1}{2}}}(B_t^u)\right)\right)$.
    \State $\theta_{t+1} \gets \mathrm{Opt}_{\theta}\!\left(\theta_t,\nabla_{\theta}L_{\mathrm{pre}}^{\mathrm{learn}}\right)$.
    \State $\phi_{t+1} \gets \phi_{t+\frac{1}{2}}$.
\EndFor
\end{algorithmic}
\end{algorithm}

\cref{alg:vtraining} gives the generic \name procedure. Each iteration has two conceptually separate updates. First, the feedback batch is used to update the task designer through the value objective. Second, the learner is updated only on the resulting pretraining loss over the unlabeled batch. The regularizer $R(\phi)$ is optional and instantiation-specific. It can restrict the designer to stay close to a base pretraining recipe, for example by limiting the amount of probability mass moved away from the true next token \citep{cohen2024idk} or by enforcing mask sparsity and spatial smoothness \citep{shi2022adios}. These constraints prevent the designer from constructing arbitrary adversarial tasks and keep the learner update anchored to the unlabeled example.

\subsection{Information bottlenecks}
\label{app:invariants}

The algorithm is defined by four information bottlenecks. (1) Feedback labels appear only in the evaluator. The feedback set $\mathcal{D}_{\mathrm{fb}}$ is used to compute $L_{\mathrm{down}}$ and $g_{\mathrm{down}}$. These labels or rewards are not inserted into the learner's pretraining batch. (2) The downstream gradient is not a learner update. Although $g_{\mathrm{down}}$ is a gradient with respect to learner parameters, it is never passed to the learner optimizer. It is detached and used only as an evaluator vector for updating $\phi$. (3) The learner update is self-supervised. The learner step minimizes $L_{\mathrm{pre}}^{\mathrm{learn}}$ on unlabeled data. The targets, masks, or views are produced by the designer but detached before the learner update. Thus the learner receives no supervised downstream loss. (4) Feedback reaches the learner only through task construction. The only path from downstream feedback to $\theta$ is
\begin{equation}
    \mathcal{D}_{\mathrm{fb}}
    \rightarrow
    g_{\mathrm{down}}
    \rightarrow
    \phi
    \rightarrow
    c_{\phi}(B_u)
    \rightarrow
    \nabla_{\theta}L_{\mathrm{pre}}.
\end{equation}

In practice, computing the alignment over all learner parameters can be expensive. We therefore allow the value to be computed on a subset of learner parameters $\mathcal{S}$, such as adapters, the last transformer blocks, or task-relevant projection layers:
\begin{equation}
    V_{\mathcal{S}}(\phi;\theta,B_u,B_f)
    =
    g_{\mathrm{down},\mathcal{S}}^{\top}
    g_{\mathrm{pre},\mathcal{S}}.
\end{equation}
When $\mathcal{S}$ is the full parameter set, this recovers the original $V$. When $\mathcal{S}$ is restricted, the objective is a computationally cheaper estimator of the same alignment principle. The derivation above is written for an SGD-style step to expose the basic mechanism. For adaptive optimizers, the same Taylor argument applies to the actual optimizer-induced update direction. Our implementation uses gradient alignment as a lightweight surrogate under a fixed learner optimizer and validates this surrogate empirically with a one-step probe

\section{One-step Value Objective}
\label{app:value-objective}

\subsection{Derivation}
\label{app:value-derivation}

At a training iteration, let $B^u\subset\mathcal D_u$ be an unlabeled pretraining batch and $B^f\subset\mathcal D_{\rm fb}$ be a feedback batch. The designer induces a candidate pretraining gradient
\begin{equation}
    g_{\rm pre}(\theta;\phi,B^u)
    =
    \nabla_\theta L_{\rm pre}\bigl(\theta;c_\phi(B^u)\bigr),
    \label{eq:app-gpre}
\end{equation}
and the feedback batch defines an evaluator gradient
\begin{equation}
    g_{\rm down}(\theta;B^f)
    =
    \nabla_\theta L_{\rm down}(\theta;B^f).
    \label{eq:app-gdown}
\end{equation}
For the candidate learner update
\begin{equation}
    \theta^+=\theta-\eta g_{\rm pre}(\theta;\phi,B^u),
\end{equation}
a first-order expansion of the feedback loss gives
\begin{equation}
    L_{\rm down}(\theta^+;B^f)
    \approx
    L_{\rm down}(\theta;B^f)
    -
    \eta\,
    g_{\rm down}(\theta;B^f)^\top
    g_{\rm pre}(\theta;\phi,B^u).
    \label{eq:app-first-order}
\end{equation}
Thus the value score
\begin{equation}
    V(\phi;\theta,B^u,B^f)
    =
    g_{\rm down}(\theta;B^f)^\top g_{\rm pre}(\theta;\phi,B^u)
    \label{eq:app-value}
\end{equation}
estimates the predicted one-step decrease in downstream loss induced by the proposed self-supervised update. \name trains the designer by minimizing
\begin{equation}
    L_{\rm meta}(\phi)=-V(\phi;\theta,B^u,B^f)+\lambda R(\phi),
    \label{eq:app-meta-loss}
\end{equation}
where $R$ is an optional regularizer that keeps the constructed task close to the base pretraining recipe.

When updating $\phi$, $g_{\rm down}$ is treated as a detached evaluator vector:
\begin{equation}
    \nabla_\phi L_{\rm meta}(\phi)
    =
    -\nabla_\phi
    \left[
        \operatorname{sg}(g_{\rm down}(\theta;B^f))^\top
        \nabla_\theta L_{\rm pre}(\theta;c_\phi(B^u))
    \right]
    +\lambda\nabla_\phi R(\phi).
    \label{eq:app-meta-gradient}
\end{equation}
This is a mixed Hessian-vector product. The designer is updated according to how its construction changes the learner gradient, as judged by the downstream evaluator gradient.

\subsection{Parameter subset used for alignment}
\label{app:subset}

Computing the alignment over all learner parameters can be expensive. We therefore compute the value objective on a subset $S$ of learner parameters:
\begin{equation}
    V_S(\phi;\theta,B^u,B^f)
    =
    g_{{\rm down},S}(\theta;B^f)^\top
    g_{{\rm pre},S}(\theta;\phi,B^u).
    \label{eq:app-subset-value}
\end{equation}
In language experiments, $S$ is chosen from the final learner blocks or other task-relevant parameters. In the current profiled meta run, the alignment uses the last two learner layers. In vision experiments, $S$ is chosen from the final backbone blocks used by the downstream evaluator heads. This reduces the cost of the mixed derivative while keeping the value signal tied to features used by the feedback tasks.

\subsection{One-step justification and bounds}\label{app:proofs}

We now state the local guarantee underlying the value objective. The result is not a long-horizon convergence theorem. Instead, it shows that the alignment objective maximizes a first-order lower bound on one-step downstream improvement.
\begin{proposition}[1-step downstream improvement lower-bound]\label{prop:value_descent}
Let $L_{\mathrm{down}}$ be $L$-smooth in $\theta$. Let
\begin{equation}
    \theta^{+}
    =
    \theta
    -
    \eta g_{\mathrm{pre}}(\theta;\phi)
\end{equation}
for step size $\eta>0$, and define
\begin{equation}
    V(\phi;\theta)
    =
    g_{\mathrm{down}}(\theta)^{\top}
    g_{\mathrm{pre}}(\theta;\phi),
    \qquad
    g_{\mathrm{down}}(\theta)=\nabla_{\theta}L_{\mathrm{down}}(\theta).
\end{equation}
Then
\begin{equation}\label{eq:value-descent-ineq}
    L_{\mathrm{down}}(\theta)
    -
    L_{\mathrm{down}}(\theta^{+})
    \ge
    \eta V(\phi;\theta)
    -
    \frac{L\eta^2}{2}
    \left\|g_{\mathrm{pre}}(\theta;\phi)\right\|_2^2.
\end{equation}
\end{proposition}

\begin{proof}[Proof of \cref{prop:value_descent}]
By $L$-smoothness,
\begin{equation}
    L_{\mathrm{down}}(\theta^{+})
    \le
    L_{\mathrm{down}}(\theta)
    +
    \nabla_{\theta}L_{\mathrm{down}}(\theta)^{\top}
    (\theta^{+}-\theta)
    +
    \frac{L}{2}
    \|\theta^{+}-\theta\|_2^2.
\end{equation}
Substituting $\theta^{+}-\theta=-\eta g_{\mathrm{pre}}(\theta;\phi)$ gives
\begin{equation}
    L_{\mathrm{down}}(\theta^{+})
    \le
    L_{\mathrm{down}}(\theta)
    -
    \eta g_{\mathrm{down}}(\theta)^{\top}g_{\mathrm{pre}}(\theta;\phi)
    +
    \frac{L\eta^2}{2}
    \|g_{\mathrm{pre}}(\theta;\phi)\|_2^2.
\end{equation}
Rearranging yields \cref{eq:value-descent-ineq}.
\end{proof}

The proposition explains the role of $V$. For sufficiently small learner steps, or when the pretraining update norm is controlled, increasing $g_{\mathrm{down}}^{\top}g_{\mathrm{pre}}$ increases a certified lower bound on one-step downstream loss decrease. Equivalently, define the one-step bilevel objective
\begin{equation}
    J(\phi;\theta)
    =
    L_{\mathrm{down}}
    \left(
        \theta
        -
        \eta\nabla_{\theta}L_{\mathrm{pre}}(\theta;\phi)
    \right).
\end{equation}
A Taylor expansion gives $J(\phi;\theta)=L_{\mathrm{down}}(\theta)-\eta V(\phi;\theta)+O(\eta^2)$. Thus maximizing $V$ is equivalent to minimizing the first-order approximation of the one-step downstream objective.

This local result is deliberately modest. It does not claim that every high-value step guarantees monotonic downstream improvement over a long training trajectory. Stochastic gradients, optimizer state, distribution shift between feedback and evaluation, and interactions across steps can all affect the final trajectory. The purpose of the value objective is to provide a scalable online signal for task design. \cref{sec:experiments} evaluates whether it improves downstream performance over full continued-pretraining runs.

\subsection{Relation to the actual optimizer}
\label{app:optimizer-note}

The main text and pseudocode use gradient descent for clarity. If the realized learner optimizer update is $d_t(\phi)=\theta^+(\phi)-\theta_t$, the exact first-order value of that update is
\begin{equation}
    -g_{\rm down}(\theta_t)^\top d_t(\phi).
\end{equation}
For a preconditioned step $d_t(\phi)\approx -\eta_t P_t g_{\rm pre}(\theta_t;\phi)$, this value becomes $g_{\rm down}(\theta_t)^\top P_t g_{\rm pre}(\theta_t;\phi)$. Our implementation uses the unpreconditioned alignment $g_{\rm down}^\top g_{\rm pre}$ as a lightweight surrogate under a fixed learner optimizer and schedule. The one-step probe in \cref{app:one-step-probe} empirically checks that this surrogate is positively correlated with realized downstream loss decrease in the language setting.

%% file: section/app_experiments.tex
\section{Additional Language Results}
\label{app:additional-language}

\subsection{One-step value probe}
\label{app:one-step-probe}

The value objective predicts the one-step downstream loss decrease as $\eta g_{\rm down}^\top g_{\rm pre}$. To test whether this estimate is informative, we compute the predicted improvement on held-out GSM8K probe batches and compare it with the realized decrease in probe loss after a gradient-descent update on $g_{\rm pre}$. Across probe measurements, the predicted and realized improvements have Pearson correlation $r=0.657$. The correlation is not expected to be perfect because the estimate is local, minibatch-based, and computed under a fixed surrogate for the realized optimizer step. Its positive value supports using the alignment score as an online training signal for the designer.

\subsection{Decontamination}
\label{app:decontamination}

To test whether the language gains are explained by near-duplicates in the continued-pretraining stream, we decontaminate NuminaMath-CoT by removing examples similar to GSM8K and MATH. We use MinHash LSH and $n$-gram Jaccard similarity to identify candidate overlaps, then retrain the Qwen1.5-4B baseline and \name models under the same wall-clock budget. \cref{tab:app-decontamination} shows that \name remains above the baseline after decontamination, though with a smaller margin than on the original stream.

\begin{table}[h]
\centering
\small
\begin{tabular}{lrr}
\toprule
Unlabeled stream & Baseline & \name \\
\midrule
Original NuminaMath-CoT & 56.48 & 58.98 \\
Decontaminated NuminaMath-CoT & 56.70 & 57.50 \\
\bottomrule
\end{tabular}
\caption{GSM8K Pass@1 for Qwen1.5-4B before and after removing near-duplicates of GSM8K and MATH from the continued-pretraining stream.}
\label{tab:app-decontamination}
\end{table}

\subsection{Generalization beyond the GSM8K verifier}
\label{app:language-generalization}

\cref{tab:app-generalization} evaluates whether GSM8K feedback collapses the learner onto the verifier task. \name improves OMEGA for the 4B learner and leaves MMLU nearly unchanged for the 4B and 7B learners. The 0.5B learner drops on MMLU, indicating that small models can be more sensitive to narrow feedback. These results support the interpretation of \name as a steering mechanism rather than a uniform improvement guarantee.

\begin{table}[h]
\centering
\small
\begin{tabular}{llrr}
\toprule
Benchmark & Model & Baseline & \name \\
\midrule
OMEGA Acc. $\uparrow$ & Qwen1.5-0.5B & 0.65 & 0.65 \\
OMEGA Acc. $\uparrow$ & Qwen1.5-4B & 1.44 & 1.88 \\
OMEGA Acc. $\uparrow$ & Qwen1.5-7B & 1.52 & 1.50 \\
MMLU Acc. $\uparrow$ & Qwen1.5-0.5B & 38.08 & 35.01 \\
MMLU Acc. $\uparrow$ & Qwen1.5-4B & 53.32 & 53.51 \\
MMLU Acc. $\uparrow$ & Qwen1.5-7B & 58.81 & 58.68 \\
\bottomrule
\end{tabular}
\caption{Language generalization beyond the GSM8K feedback task. Larger learners do not collapse onto the verifier task, while the 0.5B learner shows a drop on MMLU.}
\label{tab:app-generalization}
\end{table}

\subsection{Feedback-set coverage}
\label{app:feedback-coverage}

We vary the number of GSM8K feedback examples used to compute the evaluator gradient, using 1k, 2k, and 3k examples. Increasing feedback coverage improves the stability and strength of the Qwen1.5-4B gains, with diminishing returns after a few thousand examples. This suggests that the verifier set need not be large to be useful, but its quality and coverage still affect the value signal.

\begin{figure}[h]
\centering
\includegraphics[width=\linewidth]{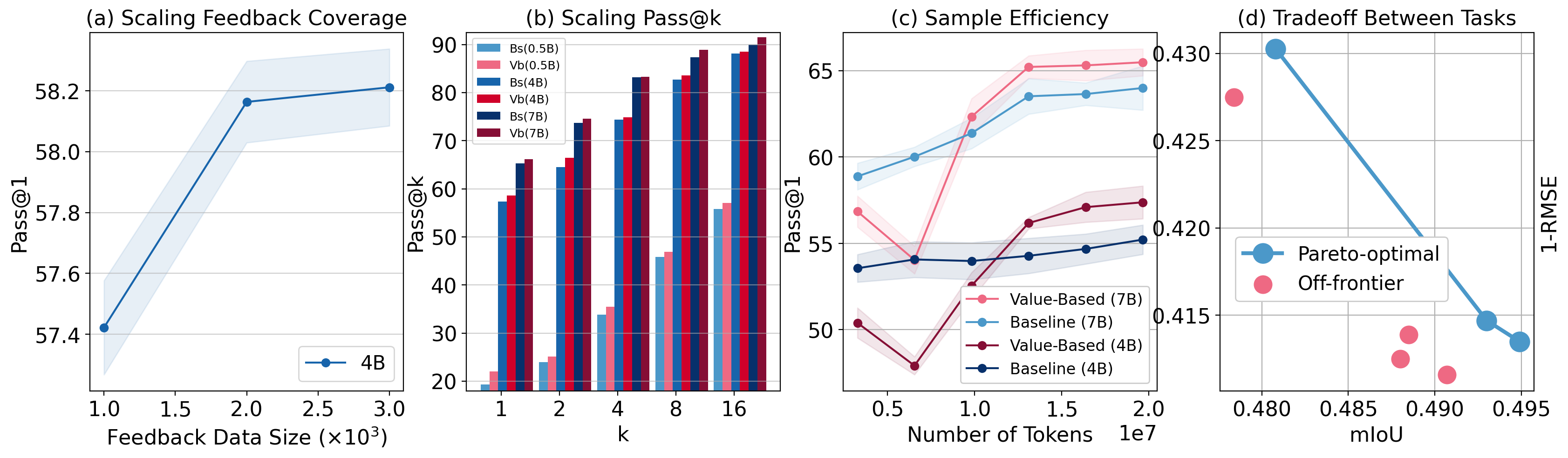}
\caption{Additional language diagnostics. (a): GSM8K Pass@1 as a function of feedback-set size. (b): Pass@$k$ across tested model sizes. (c): token-efficiency diagnostic plotting Pass@1 against unlabeled tokens processed. These curves are diagnostics, not the primary fairness criterion. (d): Tradeoff between segmentation (mIoU) and depth estimation (1-RMSE) induced by varying feedback and task-designer hyperparameters.}
\label{fig:app-language-diagnostics}
\end{figure}

\subsection{Pass@$k$}
\label{app:passk}

We evaluate Pass@$k$ for $k\in\{1,2,4,8,16\}$. \name improves Pass@$k$ across the tested $k$ values and model sizes. This indicates that the designer improves the solution distribution rather than only changing greedy decoding behavior. The result is consistent with the interpretation that value-guided targets modify reasoning-relevant update directions during continued pretraining.

\subsection{Token-efficiency diagnostic}
\label{app:token-efficiency}

The main comparisons are wall-clock matched. We additionally plot GSM8K Pass@1 as a function of unlabeled tokens processed to diagnose update quality. This diagnostic asks whether a \name update carries more downstream value per unlabeled token once the designer begins steering. In the Qwen1.5-4B setting, \name reaches 56.18 Pass@1 after 400 learner steps, approximately $1.3\times10^7$ unlabeled tokens, while the baseline requires roughly $10^3$ steps to reach comparable accuracy. The curve shows an early transient dip before the designer stabilizes; this is mitigated by a burn-in schedule that delays designer updates. This analysis is diagnostic and does not replace the wall-clock-matched results.

\subsection{Additional computation cost comparison}

\begin{table}
\centering
\small
\begin{tabular}{lccc}
\toprule
\textbf{Method} & \textbf{MATH Pass@1 $\uparrow$} &
\textbf{Hardware / time} & \textbf{Relative cost} \\
\midrule
Baseline & $52.48$ & 10 min on RTX 6000 Ada & $1\times$ \\
\name & $\mathbf{53.92}$ & 10 min on RTX 6000 Ada & $1\times$ \\
SFT+GRPO & $53.8$ & 16 hrs on H100 & $15$--$25\times$ \\
\bottomrule
\end{tabular}
\caption{Direct-vs-indirect feedback diagnostic. GRPO applies downstream reward directly to the learner; \name applies downstream feedback only to the task designer. The comparison is diagnostic, not a claim that indirect feedback replaces post-training.}
\label{tab:direct-overhead}
\end{table}

\cref{tab:direct-overhead} gives a MATH \citep{hendrycks2021math} diagnostic: \name reaches comparable Pass@1 to an SFT+GRPO pipeline in the reported setting, while using the less training-time budget. Because the stages, hardware, and optimization pipelines differ, this comparison should be read only as evidence that some downstream value can be injected before direct reward optimization.

%% file: section/app_relatedwork.tex
\section{Related Work}
\label{app:related-work}

\textbf{Positioning.}
We study \emph{controlled pretraining} under a fixed unlabeled stream and learner update budget. A small feedback set of verifiable downstream tasks provides verified goal information, but it is used only to train a lightweight controller that reshapes the \emph{pretraining target} (or views). The foundation model is \emph{never} updated on downstream labels. This differs from most label-efficient paradigms, which improve performance by creating labels or pseudo-labels and then training the main model on them.

\textbf{Post-training injects direction late.} Supervised fine-tuning and preference optimization steer models by directly updating the foundation model on labeled examples or preferences \citep{christiano2017deep,ouyang2022training,rafailov2023direct}. These methods are highly effective, but they operate after proxy pretraining has already shaped the representation space. Our approach is complementary: we inject goal information \emph{during} pretraining by shaping the unlabeled training signal rather than updating the learner on downstream labels.

\textbf{Weak/semi-supervision: scalable supervision by producing labels, not by steering pretraining updates.} A broad literature improves \emph{supervision scalability} by learning from imperfect labels or by manufacturing labels from weak sources, spanning weak supervision and data programming \citep{ratner2017snorkel,bach2017data}, distant supervision \citep{mintz2009distant}, semi-supervised learning \citep{sohn2020fixmatch}, robust learning from noisy labels \citep{song2020noisylabels}, and more recently weak-to-strong generalization as a way to elicit strong capabilities from weak supervision \citep{burns2023wsg}. Across these settings, progress typically comes from generating (pseudo-)labels and then training the \emph{main model} on task-defined targets, often with repeated inference or teacher--student refinement that is not compute-matched to pretraining-scale update budgets. Our method can be viewed as a task-agnostic \emph{pretraining analogue} of weak-to-strong generalization: a small feedback set of verifiable downstream tasks provides weak but reliable goal information, yet the foundation model is never trained on downstream labels; instead, the feedback trains a lightweight controller that reshapes the \emph{self-supervised} target/views so that each unlabeled gradient step has higher downstream value.

\textbf{Directing pretraining without step-level downstream feedback: proxy objectives and view design.} Most improvements to foundation-model pretraining change the proxy objective or the view/augmentation pipeline while keeping the training signal fixed: in language this includes next-token-based variants and domain-shaped objectives specified \emph{a priori} \citep{gpt3,zhang2019pegasus,bachmann2025pitfalls,shao2025continuous}, while in vision SSL many methods learn from global semantics via contrastive/joint-embedding objectives \citep{chen2020simclr,grill2020byol,caron2021emerging} and others inject spatial structure through handcrafted augmentations or predictive objectives such as masked modeling and JEPA-style prediction \citep{he2022mae,assran2023self}. These approaches can yield strong representations, but the \emph{direction} they impose is largely static: the target construction does not adapt online to what a downstream verifier says is valuable for the current model and example \citep{shi2022adios,bandara2023cvpr}. In contrast, value-based pretraining introduces a control loop that uses a small feedback set of verifiable downstream tasks to \emph{modulate the pretraining target/views} so that each unlabeled update aligns with downstream improvement, directly addressing the value-per-step and feedback-efficiency pressures highlighted in our introduction.

\textbf{Bilevel optimization and influence.} Downstream-aware task design naturally leads to bilevel optimization and unrolled differentiation through training \citep{maclaurin2015gradient,franceschi2018bilevel}, which is costly at pretraining horizons. To our knowledge, existing work has not optimized both pretraining tasks and SSL augmentations in a bilevel optimization \citep{reed2021self}; the closest approaches use a coordinate-descent-like step-wise optimization \cite{you2021graph,you2022bringing,jin2022automated}. 
We circumvent the computational challenges of this bilevel optimization using influence-style methods that estimate the effect of training updates on downstream loss from gradients \citep{koh2017understanding,pruthi2020estimating}.
We build on these approximations but apply them to \emph{target/view construction during pretraining}: a controller learns to reshape the unlabeled supervision signal so that each proxy update aligns with downstream improvement.

%% file: section/app_multitask.tex
\section{Multitask Feedback Data Construction}
\label{app:multitask-data}

\paragraph{Goal.}
The multitask language experiment is designed to test whether the same indirect feedback channel can steer continued pretraining toward multiple capability domains in a single run. We consider three domains: mathematical reasoning, code generation, and general instruction following / world knowledge. The key design constraint is that feedback should provide a clean value signal for the task designer, while the learner still trains on a mixed continued-pretraining stream.

\paragraph{Pretraining mixture.}
The final multitask continued-pretraining corpus contains both raw domain text and high-quality instruction-format data. The raw component provides broad next-token prediction coverage; the high-quality component anchors the stream in task-relevant
formats. The three domains are:

\begin{itemize}
    \item \textbf{Math:} OpenWebMath \citep{paster2023openwebmath} as raw mathematical text, and MetaMathQA \citep{yu2023metamath} as
    high-quality math instruction data.
    \item \textbf{Code:} codeparrot-clean \citep{codeparrot_clean2022} as raw code-domain text, and Magicoder OSS-Instruct-75K \citep{wei2024magicoder} as high-quality code instruction data.
    \item \textbf{General:} C4 English \citep{2020t5} as raw web text, and Alpaca \citep{taori2023alpaca} as high-quality
    general instruction data.
\end{itemize}

The pretraining mixture assigns total probability mass \(0.7\) to raw data and \(0.3\) to high-quality instruction data. Within each group, probability mass is balanced equally across the three domains. Thus each raw domain receives mass \(0.7/3\), and each high-quality domain receives mass \(0.3/3\). Sampling is weighted rather than performed by simple concatenation, so large corpora such as C4 do not dominate smaller instruction datasets.

\paragraph{Why mix raw and high-quality data?}
We found that instruction-only mixtures make the experiment less diagnostic. If the pretraining stream consists almost entirely of clean instruction pairs (which is often infeasible at practical pretraining scale), the standard baseline can learn much of the task format directly from the pretraining data. In that case, improvements may reflect direct exposure to instruction data rather than feedback-controlled task construction. Conversely, very small or poorly balanced mixtures can over-repeat small domains, especially code, and create domain-specific overfitting. The final raw/high-quality mixture therefore separates two roles: raw text supplies broad domain coverage, while high-quality examples keep the stream aligned with the capabilities evaluated downstream.

\paragraph{Sampling caps.}
Before tokenization and chunking, we cap the number of source items drawn from each domain. In the full multitask builder, the default caps are 50k raw source
documents per domain and 5k high-quality examples per domain for the pretraining portion. Because raw documents are chunked into fixed-length token sequences after tokenization, the number of resulting training instances can be substantially larger than the number of original source documents. Smaller pilot runs use the same construction logic with reduced source counts.

\paragraph{Sequence construction.}
All examples are tokenized with maximum sequence length 512. Raw-text examples are treated as standard next-token prediction data: an end-of-sequence token is appended, the token stream is chunked into contiguous sequences of length 512, and the learner predicts every token in the chunk. Very short chunks are discarded.

High-quality instruction examples are converted to a shared instruction-response format. Each example contains an instruction, an optional input field, and a response. The learner loss is masked on the prompt tokens and applied only to the response tokens. This formatting is used for MetaMathQA, Magicoder, and Alpaca so that the three high-quality domains share a common surface structure.

\paragraph{Feedback data.}
The feedback batches use high-quality instruction-format data only. We use MetaMathQA for math feedback, Magicoder OSS-Instruct-75K for code feedback, and Alpaca for general instruction feedback. These feedback pools are balanced equally
across the three domains. For a feedback task \(j\), let \(g_j\) denote the evaluator gradient computed from the corresponding feedback batch. The multitask
value signal is
\begin{equation}
    g_{\rm down}
    =
    \sum_{j=1}^J \omega_j g_j ,
\end{equation}
where \(\omega_j\) controls the relative weight of task \(j\). In the reported multitask run, feedback batches are domain-balanced; the resulting combined
gradient trains the task designer to construct pretraining updates that align with multiple downstream-relevant directions.

\paragraph{Separation of roles.}
The same dataset source can play different roles depending on how it is used. Examples in the continued-pretraining mixture are learner data: they define next-token prediction updates. Feedback examples are evaluator data: they define \(g_j\) for the task designer. They are not inserted as learner updates during the meta step. Thus the multitask experiment preserves the indirect-feedback channel:
the learner is trained on the continued-pretraining stream, while the feedback sets train only the task designer.

\paragraph{Pilot mixtures.}
We tested two earlier data constructions before using the final raw/high-quality mixture. The first was a MetaMathQA-heavy blend with small code and MMLU-style
components. This made the math pretraining stream too close to GSM8K/MATH-style evaluation and required substantial repetition of small code subsets. The second was an equal-weight instruction-only mixture using filtered math instruction data, Magicoder, and Alpaca. This removed the most obvious over-repetition issue, but the setting remained too clean: both the baseline and \name could learn strongly from direct instruction-format continuation alone. These pilot results motivated the final mixture, which combines broad raw-domain coverage with a smaller amount of high-quality instruction data.

\paragraph{Evaluation.}
The multitask run is evaluated on held-out benchmarks that probe the three target capability domains: GSM8K for mathematical reasoning, MMBP/MBPP-style code
generation, and MMLU for broad knowledge and instruction following. These benchmarks are used for reporting downstream performance, not as learner-update targets. The purpose of the multitask experiment is not to show that all metrics improve uniformly, but to test whether a combined feedback gradient can steer one continued-pretraining run toward several capabilities under a fixed compute budget.

%% file: section/app_main_lang.tex
\section{Main Language Experiment Setup}
\label{app:main-language-setup}

\paragraph{Purpose.}
The main language experiments test whether a small downstream verifier can steer continued language-model pretraining without directly training the learner on the verifier examples. The target capability is mathematical reasoning. The learner is
continued on a math-oriented pretraining stream, while a small GSM8K feedback set is used only to train the task designer through the gradient-alignment objective. The reported metric is GSM8K test Pass@1.

\subsection{Single-task math continued pretraining}
\label{app:single-task-language}

\paragraph{Learners.}
The main single-task language results use Qwen-family \citep{qwen1.5,qwen2.5} causal language models as learners. The reported runs include Qwen1.5 learners at multiple scales and a Qwen2.5-0.5B learner. Unless otherwise stated, each method starts from the same initial checkpoint for a given model size and uses the same learner optimizer, learning-rate schedule, numerical precision, sequence length, and wall-clock training budget.

\paragraph{Unlabeled pretraining stream.}
The single-task math runs use NuminaMath-CoT \citep{numina_math_datasets} as the continued-pretraining stream. Each example is formatted as a problem followed by a solution. Training uses causal language modeling on the solution span: prompt tokens are masked in the loss, and the learner predicts only answer or solution tokens. Both the baseline and \name use the same pretraining stream and the same loss mask.

\paragraph{Baseline.}
The baseline is standard continued next-token pretraining on the math stream. For a token position \(t\) with true next token \(y_t\), the learner is trained with a one-hot target \(\delta_{y_t}\):
\begin{equation}
    L_{\rm base}(\theta)
    =
    \frac{1}{|B^u|}
    \sum_{x\in B^u}
    \frac{1}{|I(x)|}
    \sum_{t\in I(x)}
    {\rm CE}\!\left(\delta_{y_t},p_\theta(\cdot\mid x_{<t})\right),
\end{equation}
where \(B^u\) is an unlabeled pretraining batch, $\rm CE$ denotes the cross-entropy loss, and \(I(x)\) denotes the solution-token positions included in the learner loss.

\paragraph{Feedback set.}
The feedback set contains 1,024 GSM8K training examples. These examples define the downstream evaluator loss used to compute \(g_{\rm down}\), but they are not inserted into the learner's pretraining stream and are not used as supervised learner updates. For a feedback batch \(B^f\), the evaluator loss is
\begin{equation}
    L_{\rm down}(\theta;B^f)
    =
    \frac{1}{|B^f|}
    \sum_{(u,a)\in B^f}
    \frac{1}{|A(u,a)|}
    \sum_{t\in A(u,a)}
    -\log p_\theta(a_t\mid u,a_{<t}),
\end{equation}
where \(u\) is the problem statement, \(a\) is the verified solution, and \(A(u,a)\) denotes answer-token positions. The resulting gradient
\(g_{\rm down}=\nabla_\theta L_{\rm down}(\theta;B^f)\) is detached and used only as an evaluator vector for the task designer.

\paragraph{\name objective.}
\name keeps the same text stream and causal-LM learner but replaces the fixed one-hot next-token target with a designer-shaped soft target over a small candidate set. For each eligible token position \(t\), we form
\begin{equation}
    C_t
    =
    \{y_t\}
    \cup
    {\rm TopK}_{K-1}\!\left(
        {\rm sg}\!\left(p_\theta(\cdot\mid x_{<t})\right)
        \setminus \{y_t\}
    \right),
\end{equation}
so the true next token is always included and the remaining candidates are the learner's high-probability alternatives. The task designer outputs a distribution \(r_{\phi,t}\in\Delta(C_t)\) and a token-dependent mixing coefficient \(\alpha_{\phi,t}\in[0,\alpha_{\max}]\). The learner target is
\begin{equation}
    q_{\phi,t}(v)
    =
    (1-\alpha_{\phi,t})\mathbf 1[v=y_t]
    +
    \alpha_{\phi,t}r_{\phi,t}(v),
    \qquad v\in C_t,
\end{equation}
and \(q_{\phi,t}(v)=0\) for \(v\notin C_t\). Setting
\(\alpha_{\phi,t}=0\) recovers the baseline one-hot objective.

During a meta update (designer update), the designer is trained to make the induced pretraining gradient align with the downstream evaluator gradient. Let
\begin{equation}
    g_{\rm pre,S}
    =
    \nabla_{\theta_S}
    L_{\rm pre}^{\rm LM}(\theta;\phi,B^u),
    \qquad
    g_{\rm down,S}
    =
    {\rm sg}\!\left(
    \nabla_{\theta_S}L_{\rm down}(\theta;B^f)
    \right),
\end{equation}
where \(S\) is the subset of learner parameters used for alignment. The implementation maximizes gradient alignment, using either the dot product from the main derivation or its cosine-normalized version:
\begin{equation}
    L_{\rm meta}(\phi)
    =
    -{\rm Align}(g_{\rm down,S},g_{\rm pre,S}).
\end{equation}
After the designer update, the soft targets are recomputed and detached. The learner update is then an ordinary causal-LM pretraining update:
\begin{equation}
    g^{\rm learn}
    =
    \nabla_\theta
    L_{\rm pre}^{\rm LM}
    \bigl(\theta;{\rm sg}(q_\phi),B^u\bigr),
    \qquad
    \theta^+
    =
    \theta-\eta g^{\rm learn}.
\end{equation}
Thus GSM8K feedback affects the learner only by changing the self-supervised next-token targets on NuminaMath-CoT examples. No GSM8K supervised gradient is applied to the learner in our main experiments (\cref{tab:main-results}).

\paragraph{Evaluation.}
We evaluate on the GSM8K test set using the same decoding and answer-extraction protocol for all methods in a comparison. Main results report Pass@1. Additional Pass@\(k\), feedback-coverage, and token-efficiency diagnostics are reported in separate appendix sections.

\subsection{Language implementation branches}
\label{app:language-branches}

We used two related language implementations for different experiments. They share the same conceptual student--designer structure, but differ in learner parameterization, data regime, and alignment parameters. We report them separately to avoid conflating parameter-efficient adaptation with full-parameter continued pretraining.

\paragraph{LoRA reference implementation.}
The compact reference implementation uses Qwen2.5-0.5B as the learner and trains
LoRA adapters rather than all model parameters. The LoRA configuration is
\[
    r=8,\qquad \alpha_{\rm LoRA}=16,\qquad {\rm dropout}=0.05.
\]
Adapters are applied to the standard attention and MLP projections:
\texttt{q\_proj}, \texttt{k\_proj}, \texttt{v\_proj}, \texttt{o\_proj},
\texttt{gate\_proj}, \texttt{up\_proj}, and \texttt{down\_proj}. This branch is therefore a parameter-efficient language adaptation setting.

The LoRA reference branch uses a multitask instruction-style pretraining stream with MathInstruct, Magicoder OSS-Instruct-75K, and Alpaca. MathInstruct is filtered to remove sources derived from GSM8K and MATH. All examples are converted to a shared causal-LM format with prompt masking, so only answer tokens contribute to the learner loss.

The task designer is a compact decoder-only TopKAugmentor. It scores only the student learner's top-\(K\) next-token candidates rather than the full vocabulary. It outputs both the top-\(K\) distribution \(r_{\phi,t}\) and the smoothing gate
\(\alpha_{\phi,t}\). Gradient alignment is computed on LoRA parameters from the last two transformer layers. The default settings in this branch are:
\[
\begin{array}{ll}
\text{student learning rate} & 2\times 10^{-4},\\
\text{designer learning rate} & 1\times 10^{-4},\\
\text{batch size} & 4,\\
\text{gradient accumulation} & 8,\\
\text{maximum steps} & 2000,\\
\text{pretraining sequence length} & 1024,\\
\text{downstream sequence length} & 512,\\
\text{meta sequence length} & 256,\\
K & 64,\\
\alpha_{\max} & 0.5,\\
\text{meta-update frequency} & \text{every 8 learner steps},\\
\text{alignment subset} & \text{last 2 LoRA layers}.
\end{array}
\]

\paragraph{Full-parameter implementation.}
The full-parameter implementation uses Qwen2.5-0.5B with all learner parameters trainable. This branch is a direct full continued-pretraining implementation of the same feedback-controlled task-design idea. The baseline trains the learner
with ordinary causal language modeling on the selected pretraining stream. Raw text uses standard next-token prediction, while instruction-format examples mask prompt tokens and train only on response tokens.

The default full-parameter baseline uses AdamW with bf16 learner precision and
the following settings:
\[
\begin{array}{ll}
\text{learning rate} & 1.5\times 10^{-5},\\
\text{batch size} & 4,\\
\text{gradient accumulation} & 32,\\
\text{effective batch size} & 128,\\
\text{epochs} & 2,\\
\text{warmup ratio} & 0.03,\\
\text{sequence length} & 512,\\
\text{weight decay} & 0,\\
\text{max gradient norm} & 1.0,\\
\text{learner precision} & \text{bf16}.
\end{array}
\]

The corresponding \name version uses the same learner and pretraining stream, but adds a small TopKAugmentor. Unless otherwise stated, the augmentor has hidden size 256, 6 layers, 4 attention heads, top-\(K=64\), and \(\alpha_{\max}=0.5\). The augmentor is trained in fp32 even when the learner is
trained in bf16. Alignment is computed over a selected subset of full learner parameters, usually the last few transformer layers, controlled by
\texttt{align\_last\_n\_layers}.

\paragraph{Domain-matched alignment for multitask language runs.}
In multitask language runs, we compute gradient alignment within matching domains rather than pooling all tasks into a single undifferentiated feedback batch. For a domain \(d\in\{\text{math},\text{code},\text{general}\}\), the method pairs a same-domain pretraining batch with a same-domain feedback batch:
\begin{equation}
    B^u_d \leftrightarrow B^f_d.
\end{equation}
It then computes
\begin{equation}
    g_{{\rm pre},d}
    =
    \nabla_{\theta_S}
    L_{\rm pre}(\theta;\phi,B^u_d),
    \qquad
    g_{{\rm down},d}
    =
    {\rm sg}\!\left(
    \nabla_{\theta_S}L_{\rm down}(\theta;B^f_d)
    \right),
\end{equation}
and trains the designer with the mean alignment objective
\begin{equation}
    L_{\rm meta}(\phi)
    =
    -
    \frac{1}{|\mathcal D|}
    \sum_{d\in\mathcal D}
    {\rm Align}(g_{{\rm down},d},g_{{\rm pre},d}).
\end{equation}
This domain-matched design avoids a failure mode in which a single combined feedback gradient pulls the designer toward a direction inconsistent with the
current pretraining domain.

\paragraph{Feedback formatting variants.}
For multitask language experiments, the default feedback pools are MetaMathQA for math, Magicoder for code, and Alpaca for general instruction
following. These feedback examples are formatted consistently with the instruction-style pretraining examples. We also implemented an evaluation-aligned
feedback variant, in which feedback examples are reformatted to more closely match downstream benchmark prompts: math feedback is written in a GSM8K-style format, code feedback in an MBPP-style format with markdown fences removed, and knowledge feedback in an MMLU multiple-choice format. This variant is treated as a separate experimental branch rather than the universal default.

\paragraph{Separation of learner and feedback data.}
Across both implementation branches, the key invariant is unchanged. Pretraining examples define learner updates. Feedback examples define evaluator gradients for the task designer. Even when a feedback dataset has the same source name as a
high-quality pretraining dataset, its role is different: feedback batches are not inserted as learner updates during the meta step. The only learner update remains a causal-LM loss on the current pretraining batch with detached designer-shaped targets.

%% file: section/app_main_vision.tex
\section{Main Vision Experiment Setup}
\label{app:main-vision-setup}

\paragraph{Purpose.}
The vision experiments test whether downstream dense-prediction feedback can steer continued self-supervised pretraining without directly training the visual backbone on dense labels. The learner is a DINOv3 vision transformer \citep{simeoni2025dinov3} continued on unlabeled ImageNet-1K images with a DINO-style self-supervised objective. The feedback tasks are ADE20K semantic segmentation \citep{zhou2017ade20k} and NYUv2 depth estimation \citep{silberman2012nyu}. Their labels are used only to train the task designer through the value objective; the backbone learner is still updated only by a self-supervised loss on unlabeled images.

\subsection{Baseline continued DINO pretraining}
\label{app:vision-baseline}

\paragraph{Learners and unlabeled data.}
The main vision results use pretrained DINOv3 ViT backbones, including ViT-B and ViT-L (we discuss the I-JEPA setup later). Each run starts from the same pretrained checkpoint for a given backbone size and continues self-supervised training on ImageNet-1K. The same unlabeled image stream, crop configuration, learner optimizer, learning-rate schedule, teacher momentum schedule, numerical precision, and training budget are used for the baseline and \name comparison.

\paragraph{DINO-style SSL objective.}
The baseline is standard continued DINO-style self-supervised learning. The student consists of the DINOv3 backbone and a projection head. The teacher is an exponential-moving-average copy of the student. For each image, the fixed augmentation pipeline samples multiple views: two global crops and six local
crops by default, with color jitter, grayscale augmentation, Gaussian blur, and solarization \citep{caron2020multicrop,caron2021emerging}. The student is trained to match teacher predictions across non-matching views, with temperature scaling and a running teacher-output center.

Let \(V_0(x)=\{v^0_1,\ldots,v^0_M\}\) denote the views produced by the fixed DINO augmentation pipeline for an unlabeled image \(x\). The baseline minimizes
\begin{equation}
    L_{\rm ssl}^{0}(\theta;B^u)
    =
    \frac{1}{|B^u|}
    \sum_{x\in B^u}
    \ell_{\rm DINO}\!\left(\theta,\bar{\theta};V_0(x)\right),
\end{equation}
where \(\theta\) are the student parameters and \(\bar{\theta}\) are the EMA teacher parameters. The task construction rule \(c\) is fixed before training: the views are stochastic, but their sampling distribution and augmentation recipe do not depend on downstream feedback.

\paragraph{Baseline learner update.}
The baseline learner update is therefore
\begin{equation}
    g^{\rm base}_t
    =
    \nabla_\theta L_{\rm ssl}^{0}(\theta_t;B^u_t),
    \qquad
    \theta_{t+1}
    =
    \theta_t-\eta_t g^{\rm base}_t .
\end{equation}
No ADE20K or NYUv2 labels are used in the baseline continued-pretraining update.

\subsection{\name with downstream-aware learned augmentations}
\label{app:vision-meta}

\paragraph{Control surface.}
\name keeps the DINO learner, SSL objective, unlabeled stream, and optimizer fixed, but replaces part of the fixed view-construction rule with a learned instance-wise augmentation module. The designer does not predict segmentation masks or depth maps. Instead, it modifies the self-supervised views
used by DINO so that the resulting SSL gradient better aligns with dense-task feedback gradients.

\paragraph{Augmentor parameterization.}
The task designer is an image augmentor \(a_\phi\) that predicts a soft spatial mask \(\mu_\phi(v)\in[0,1]^{H\times W}\) for selected SSL crops. Two augmentor architectures are supported: a small convolutional U-Net \citep{ronneberger2015unet} mask generator and a small dit-based \citep{peebles2023dit} patchwise mask generator. In the default configuration, the augmentor is applied only to the two global SSL crops, while the local crops remain generated by the standard DINO pipeline.

Given a base crop \(v\), the learned view is formed by softly mixing the original crop with a blurred version:
\begin{equation}
    \tilde v
    =
    \mu_\phi(v)\odot v
    +
    \left(1-\mu_\phi(v)\right)\odot {\rm Blur}(v).
\end{equation}
Large mask values preserve the original pixels; small values suppress pixels by replacing them with a blurred background. For an image \(x\), let \(V_\phi(x)\) denote the resulting set of views after replacing the controlled global crops with learned views. The SSL loss under the designer is
\begin{equation}
    L_{\rm ssl}(\theta;\phi,B^u)
    =
    \frac{1}{|B^u|}
    \sum_{x\in B^u}
    \ell_{\rm DINO}\!\left(\theta,\bar{\theta};V_\phi(x)\right).
\end{equation}

\paragraph{Dense feedback tasks.}
The feedback signal comes from small labeled subsets of ADE20K and NYUv2. For the default feedback construction, we use ADE20K / SceneParse150 with 2,000 labeled training samples and 512 meta samples, and NYUv2 with 512 labeled training samples and 128 meta samples. The labeled training samples are used to
fit lightweight downstream evaluator heads, and the meta samples are used to compute downstream gradients for the task designer.

The evaluator heads are:
\[
    H_{\rm seg}
    \quad\text{for ADE20K semantic segmentation,}
    \qquad
    H_{\rm dep}
    \quad\text{for NYUv2 depth estimation.}
\]
The segmentation head is trained with pixelwise cross-entropy. The depth head is trained with an L1 loss on valid depth pixels.

\paragraph{Meta step.}
A \name meta step (task designer update step) has three stages.

First, the downstream evaluator heads are updated on small labeled training batches using frozen backbone features. This adapts the lightweight heads to the current representation without applying supervised dense-task updates to the
backbone.

Second, on separate labeled meta batches, we compute a downstream evaluator loss
\begin{equation}
    L_{\rm down}^{\rm vis}(\theta;B^f)
    =
    \alpha_{\rm seg}L_{\rm seg}
    \bigl(H_{\rm seg}(F_\theta(x)),y_{\rm seg}\bigr)
    +
    \alpha_{\rm dep}L_{\rm dep}
    \bigl(H_{\rm dep}(F_\theta(x)),y_{\rm dep}\bigr),
\end{equation}
with terms omitted when the corresponding feedback task is not used. Here \(F_\theta\) is the visual backbone, and \(\alpha_{\rm seg},\alpha_{\rm dep}\) control the relative value placed on segmentation and depth. We compute the
downstream gradient with respect to a selected subset \(S\) of backbone parameters:
\begin{equation}
    g_{\rm down,S}^{\rm vis}
    =
    {\rm sg}\!\left(
    \nabla_{\theta_S}L_{\rm down}^{\rm vis}(\theta;B^f)
    \right).
\end{equation}
In practice, \(S\) is chosen as the last \(K\) transformer blocks together with final normalization layers. This reduces the cost of the mixed second-order gradient while keeping the alignment signal tied to the representation layers used by dense prediction.

Third, a separate unlabeled ImageNet batch is passed through the learned augmentor, the DINO loss is recomputed under the learned views, and the SSL gradient is computed on the same parameter subset:
\begin{equation}
    g_{\rm ssl,S}^{\rm vis}
    =
    \nabla_{\theta_S}
    L_{\rm ssl}(\theta;\phi,B^u).
\end{equation}
The augmentor is trained to maximize alignment between the downstream dense-task gradient and the SSL gradient:
\begin{equation}
    L_{\rm meta}^{\rm vis}(\phi)
    =
    -
    \left\langle
    g_{\rm down,S}^{\rm vis},
    g_{\rm ssl,S}^{\rm vis}
    \right\rangle
    +
    \lambda_{\rm spars}R_{\rm spars}(\mu_\phi)
    +
    \lambda_{\rm tv}R_{\rm tv}(\mu_\phi).
\end{equation}
The sparsity regularizer keeps the mean mask value near a target keep ratio, and
the total-variation regularizer encourages spatially smooth masks. These
regularizers keep the learned augmentation close to plausible SSL view
construction rather than arbitrary image corruption.

\paragraph{Learner update.}
Only the augmentor is updated by \(L_{\rm meta}^{\rm vis}\). After the augmentor
update, the views are recomputed and detached. The backbone learner is then
updated by an ordinary DINO SSL loss:
\begin{equation}
    g^{\rm learn}_t
    =
    \nabla_\theta
    L_{\rm ssl}\bigl(\theta_t;{\rm sg}(V_{\phi_t}(B^u_t))\bigr),
    \qquad
    \theta_{t+1}
    =
    \theta_t-\eta_t g^{\rm learn}_t .
\end{equation}
Thus ADE20K and NYUv2 labels never appear as supervised learner targets during continued pretraining. They train only the task designer that constructs SSL views.

\paragraph{Second-order implementation detail.}
The meta update differentiates through the SSL gradient with respect to selected
backbone parameters. This requires second-order differentiation through the
backbone. To avoid unsupported double-backward paths in Flash or memory-efficient
attention kernels, the implementation uses a math-only scaled dot-product
attention backend during the relevant meta-gradient computation.

\subsection{Dense evaluation protocol}
\label{app:vision-eval}

\paragraph{ADE20K semantic segmentation.}
ADE20K evaluation uses the SceneParse150 dataset. Label \(0\) is remapped to the ignore index \(255\), and labels \(1,\ldots,150\) are remapped to \(0,\ldots,149\). Performance is reported as mean intersection-over-union (mIoU).

The evaluation code supports both a small convolutional segmentation decoder and a linear-BN segmentation head. In the sweep comparisons used for the main reported results, the evaluation protocol is fixed to a frozen-backbone linear
probe with a linear-BN head. This makes the comparison primarily a test of representation quality rather than downstream fine-tuning sensitivity.

\paragraph{NYUv2 depth estimation.}
NYUv2 evaluation uses the standard validation split. Depth values are decoded to meters, and metrics are computed on valid pixels. The evaluation protocol enables the standard Eigen crop by default. The main metric reported in the paper is RMSE; the evaluation code also computes AbsRel and \(\delta_1\).

As with segmentation, the sweep comparisons use a frozen-backbone linear probing protocol with a linear-BN depth head. Predictions are passed through a softplus nonlinearity to ensure positive depth values.

\paragraph{ImageNet linear evaluation and transfer.}
We additionally evaluate whether dense-task feedback harms global recognition. For this purpose, the main table reports ImageNet-1K linear accuracy. We also report instance-retrieval transfer on Revisited Oxford and Revisited Paris. These tasks are not used as feedback during \name and are used to check whether dense feedback collapses the representation onto segmentation or depth.

\paragraph{Evaluation hyperparameters.}
All selected checkpoints are evaluated at the last pretraining checkpoint (\texttt{eval\_pretrain\_steps=last}) with a frozen-backbone linear probing protocol (\texttt{eval\_ft\_mode=linear}) and a linear-BN head. Evaluation uses 20k optimization steps, batch size 8, bf16 precision, 1k warmup steps, weight decay \(0.05\), and gradient clipping at \(1.0\). ADE20K evaluation uses image size 512 and head learning rate \(5{\times}10^{-4}\). NYUv2 depth evaluation uses an L2 depth loss, head learning rate \(10^{-3}\), resize resolution \(480\times 640\), valid depth range \([10^{-3},10]\) meters, and the standard Eigen crop.

\subsection{Hyperparameter selection}
\label{app:vision-sweeps}

\paragraph{Fixed-budget sweep protocol.}
Vision hyperparameters are selected with fixed-budget pretraining sweeps. Each sweep trial performs 20k steps of continued DINO pretraining and then evaluates one or more saved checkpoints, with the default configuration evaluating the last
checkpoint. ADE20K-oriented sweeps optimize validation mIoU, while NYUv2-oriented sweeps optimize validation RMSE.

The main swept hyperparameters are:
\[
\begin{array}{ll}
\text{learner learning rate} & \eta,\\
\text{meta-update frequency} & r,\\
\text{meta SSL batch size} & |B^u_{\rm meta}|,\\
\text{aligned backbone blocks} & K,\\
\text{augmentor architecture} & \text{U-Net or tiny DiT},\\
\text{augmentor learning rate} & \eta_\phi,\\
\text{target mask keep ratio} & \kappa,\\
\text{sparsity regularization} & \lambda_{\rm spars},\\
\text{total-variation regularization} & \lambda_{\rm tv},\\
\text{segmentation feedback weight} & \alpha_{\rm seg},\\
\text{depth feedback weight} & \alpha_{\rm dep}.
\end{array}
\]

\paragraph{Backbone-scale adjustments.}
The sweep spaces are adjusted for backbone size to fit memory. For ViT-B, the pretraining batch size is larger; for ViT-L, the pretraining batch size and meta-SSL batch-size search space are reduced. In the larger depth sweep, the augmentor architecture is fixed to the U-Net mask generator to reduce search cost. These adjustments affect the feasible training configuration, not the definition of the feedback channel.

\paragraph{Interpretation of vision tuning.}
The vision results should be interpreted under this stated tuning protocol:
standard continued DINO pretraining and \name are compared with the same initial backbone, unlabeled ImageNet stream, SSL objective family, and evaluation protocol, while \name additionally pays the cost of learned view construction and periodic value-gradient updates. The purpose of the experiment
is not to introduce a new dense-prediction fine-tuning method, but to test whether dense feedback can improve the self-supervised pretraining updates that shape the backbone representation.

\paragraph{Selected \name configurations.}
\cref{tab:vision-selected-hparams} reports the selected \name configurations used for the main DINOv3 vision results. The target column denotes the validation metric used to select the run from the fixed-budget sweep; it does
not mean that only that feedback task is present in the meta objective. When both dense feedback loaders are enabled, the downstream evaluator gradient is formed from the weighted segmentation and depth losses with weights
\(\alpha_{\rm seg}\) and \(\alpha_{\rm dep}\).

All selected runs use 20k continued-pretraining steps, bf16 mixed precision, AdamW, weight decay \(0.04\), gradient clipping at \(1.0\), 1k warmup steps, DINO output dimension 8192, student temperature \(0.1\), teacher temperature \(0.04\), teacher momentum base \(0.996\), center momentum \(0.9\), two global crops of size 224, six local crops of size 96, and learned augmentation applied only to the global crops. The unlabeled stream is ImageNet-1K, and the feedback pools use 2,000 ADE20K training samples with 512 ADE20K meta samples and 512 NYUv2 training samples with 128 NYUv2 meta samples.

\begin{table}[t]
\centering
\scriptsize
\setlength{\tabcolsep}{3pt}
\resizebox{\textwidth}{!}{
\begin{tabular}{llrrrrrlrrrrrr}
\toprule
Target & Backbone & LR & Batch & Meta freq. & Meta SSL batch
& Aligned blocks & Augmentor & Aug. LR & Keep
& \(\lambda_{\rm spars}\) & \(\lambda_{\rm tv}\)
& \(\alpha_{\rm seg}\) & \(\alpha_{\rm dep}\) \\
\midrule
ADE20K & ViT-B
& \(4.15{\times}10^{-6}\) & 256 & 4 & 32 & 2
& U-Net & \(1.54{\times}10^{-4}\) & 0.6
& 0.131 & 0.144 & 1 & 8 \\

ADE20K & ViT-L
& \(4.50{\times}10^{-5}\) & 50 & 4 & 16 & 4
& U-Net & \(2.17{\times}10^{-4}\) & 0.5
& 0.196 & 0.0399 & 4 & 1 \\

NYUv2 & ViT-B
& \(8.58{\times}10^{-6}\) & 256 & 8 & 64 & 3
& SiT & \(2.21{\times}10^{-4}\) & 0.4
& 0.111 & 0.0417 & 16 & 2 \\

NYUv2 & ViT-L
& \(1.94{\times}10^{-6}\) & 50 & 2 & 32 & 4
& U-Net & \(8.03{\times}10^{-4}\) & 0.4
& 0.401 & 0.159 & 32 & 16 \\
\bottomrule
\end{tabular}
}
\caption{
Selected \name hyperparameters for the main DINOv3 vision runs.
Batch is the learner SSL batch size. Meta freq. is the number of learner steps
between augmentor updates. Meta SSL batch is the unlabeled ImageNet batch used
to compute the SSL gradient in the value objective. Aligned blocks denotes the
number of final ViT blocks used for gradient alignment. The target column denotes
the sweep selection metric, not the only feedback task used by the value
objective.
}
\label{tab:vision-selected-hparams}
\end{table}

%% file: section/app_declare.tex
\section{Declarations and Impacts}\label{app:declaration}

\paragraph{Use of generative AI tools.}
We used ChatGPT Image to draft the illustrative schematic in \cref{fig:illustration-of-v-pretraining} from author-written prompts. The figure contains no experimental data, generated results, or method output. We also used ChatGPT to polish some parts of the writing. All equations, labels, and scientific content were manually checked by the authors. Our use of AI tools follow the NeurIPS guidelines.

\paragraph{Broader Impacts.} As \name fundamentally enhances the capabilities of machine learning models, it inherently carries a dual-use nature. On the positive side, our approach can significantly benefit downstream applications such as creative content generation and advanced information retrieval. Conversely, we acknowledge the potential risks of malicious exploitation, particularly regarding the generation of deceptive, fake, or misleading content.

\section{Compute Resources and Total Compute}
\label{app:compute}

\paragraph{Compute workers.}
Training experiments were run on internal GPU clusters and on the DeltaAI system at the National Center for Supercomputing Applications, whose nodes use NVIDIA GH200 Grace Hopper superchips with H100 GPUs. On the internal clusters, the main training workers were NVIDIA H100 80GB and NVIDIA H200 NVL GPUs. Auxiliary evaluation, small diagnostic runs, and some post-training comparisons also used RTX 6000 Ada / RTX A6000-class GPUs with 48GB memory. We report compute in accelerator-hours, counting one hour on one GPU as one accelerator-hour. Most reported runs were single-worker training jobs; when a node exposed multiple GPUs, we count only the GPUs used by the corresponding training process.

Training jobs used CPU workers for data loading. Vision pretraining jobs used 12 dataloader workers in the selected configurations. Slurm allocations typically provided 12--16 CPU cores per GPU worker and 80--142GB of allocated host memory, with larger nodes providing substantially more physical RAM. Datasets and checkpoints were stored on shared cluster storage; individual runs wrote checkpoints, logs, and evaluation outputs, typically requiring tens of GB per run. The full project required on the order of 0.5--1TB of shared storage across datasets, checkpoints, logs, sweeps, and intermediate evaluation outputs.

\paragraph{Per-run compute.}
\cref{tab:compute-per-run} summarizes the approximate compute for the reported experimental runs. The numbers are estimates from job logs and observed throughput. They are intended to document the scale of the experiments rather than serve as exact hardware benchmarks, since wall-clock time varies with device type, dataloader throughput, cluster load, and whether evaluation is run in the
same job.

\begin{table}[t]
\centering
\small
\setlength{\tabcolsep}{4pt}
\resizebox{\textwidth}{!}{
\begin{tabular}{llrrl}
\toprule
Experiment family & Main worker & Approx. compute per run & Memory used & Notes \\
\midrule
Qwen 0.5B language continued pretraining
& 1 H100/H200
& 2--6 GPU-hours
& 30--40GB
& Baseline or \name run; GSM8K feedback for \name. \\

Qwen 4B language continued pretraining
& 1--2 H100/H200
& 10--24 GPU-hours
& 60--140GB
& Used for main GSM8K result, ablations, and decontamination controls. \\

Qwen 7B language continued pretraining
& 1--2 H100/H200
& 16--36 GPU-hours
& 80--140GB
& Used for scaling runs. \\

Language multitask run
& 1 H100/H200
& 4--12 GPU-hours
& 30--50GB
& Qwen2.5-0.5B multitask pretraining with math/code/general feedback. \\

DINOv3 ViT-B continued SSL
& 1 H100/H200
& 1--3 GPU-days
& 50--80GB
& 20k-step continued pretraining plus dense linear-probe evaluation. \\

DINOv3 ViT-L continued SSL
& 1 H100/H200
& 1--4 GPU-days
& 80--140GB
& 20k-step continued pretraining plus dense linear-probe evaluation. \\

Dense evaluation / retrieval / linear probes
& H100/H200 or RTX 6000 Ada
& 0.5--6 GPU-hours
& 20--48GB
& ADE20K, NYUv2, ImageNet linear, and retrieval evaluations. \\

Direct-feedback diagnostic
& H100 and RTX 6000 Ada
& 10 min--16 GPU-hours
& 20--80GB
& Small \name/baseline diagnostic and SFT+GRPO comparison. \\
\bottomrule
\end{tabular}
}
\caption{
Approximate compute for individual reported experimental runs. GPU-hours denote accelerator-hours. Ranges reflect differences in model size, device type, and whether evaluation is included in the same job. The reported main comparisons are
wall-clock matched within each experiment family, so \name is charged for its task-designer and value-gradient overhead.
}
\label{tab:compute-per-run}
\end{table}

\paragraph{Representative measured overhead.}
For the current language \name configuration, the task designer is a small 256-hidden-dimensional transformer with 6 layers and 8 heads, aligning the last two learner layers and performing a meta update every 8 learner optimizer steps. In a representative current run, the standard baseline used microbatch size 4, gradient accumulation 32, and sequence length 512, for \(4\times 32\times 512=65{,}536\) tokens per optimizer step. The corresponding \name run used microbatch size 4, gradient accumulation 20, and sequence length 512, for \(4\times 20\times 512=40{,}960\) tokens per optimizer step. The baseline achieved approximately 26.7k tokens/s with peak GPU memory 31,302 MiB, while \name achieved approximately 24.2k tokens/s with peak GPU memory 36,440 MiB. Thus the current \name implementation reduces token throughput by about 9\% and increases peak memory by about 16\% in this representative language setting. The shorter \name optimizer-step time is not evidence of lower cost, since each optimizer step processes fewer tokens.